\documentclass{article}
\usepackage{iclr2027_conference,times}

\usepackage{math_commands}

\usepackage{hyperref}
\usepackage{url}
\usepackage{graphicx}
\usepackage{wrapfig}
\usepackage{booktabs}
\usepackage{multirow}
\usepackage{capt-of}
\usepackage{float}
\usepackage{ragged2e}
\usepackage{tabularx}
\usepackage{caption}
\usepackage{placeins}

\title{Counterfactual Self-Evolving Agents for \\ Evidence-Grounded Reasoning}

\author{%
\begin{minipage}[t]{\dimexpr\textwidth-2\tabcolsep\relax}
\centering
Xing Han$^{1}$ \quad Yuxin Wang$^{1}$ \quad Chen Chen$^{1}$ \quad Wei Dai$^{2}$\\
Gautham Krishna Gudur$^{3}$ \quad Shijun Li$^{3}$ \quad Hsing-Huan Chung$^{3}$\\
Gregory D. Hager$^{1}$ \quad Joydeep Ghosh$^{3}$ \quad Paul Pu Liang$^{2,*}$ \quad Suchi Saria$^{1,4,*}$\\[0.6em]
\normalfont\small
$^{1}$Johns Hopkins University \quad $^{2}$MIT\\
$^{3}$University of Texas at Austin \quad $^{4}$Bayesian Health\\[0.3em]
$^{*}$Equal advising.
\end{minipage}%
}

\hypersetup{
  hidelinks,
  pdftitle={Counterfactual Self-Evolving Agents for Evidence-Grounded Reasoning},
  pdfauthor={Xing Han; Yuxin Wang; Chen Chen; Wei Dai; Gautham Krishna Gudur; Shijun Li; Hsing-Huan Chung; Gregory D. Hager; Joydeep Ghosh; Paul Pu Liang; Suchi Saria}
}

\iclrfinalcopy
\begin{document}

\maketitle
\lhead{Preprint}

\begin{abstract}
Self-play proposer--solver methods improve reasoning by generating tasks and learning from
verified solutions. However, for evidence-identifiable tasks, where case-specific
evidence and domain knowledge determine a checkable answer, self-play requires generating
plausible cases whose answers can be independently verified.
We introduce counterfactual self-evolution, which generates counterfactual context for
reconsidering the original case. A trainable Proposer constructs targeted evidence edits and
describes potential outcome changes with causal explanations. We handcraft an expert-verified
counterfactual instruction-tuning dataset to teach the Proposer to generate high-quality
counterfactuals across a broad range of action--outcome scenarios. Each counterfactual
instruction-tuning example specifies an edit within a defined category and explains its
hypothesized causal effect on the decision, teaching the Proposer to reason systematically
about what changes and why.
We instruction-tune the Proposer on these examples, then formulate a fine-tuning reward that
integrates feedback from the Solver and Verifier. Across diverse counterfactual scenarios,
this reward favors high-quality counterfactuals and warranted revisions, while penalizing
changes that overturn correct decisions. The counterfactual context aims to
correct errors and strengthen confidence in correct decisions. Accepted
counterfactuals accumulate in memory that supplies in-context evidence to the frozen Solver;
the Solver adapts through evolving context rather than weight updates. We apply the
framework to clinical reasoning, fact verification, and business reasoning. Our evaluation tracks performance over successive rounds as counterfactual memory
grows, including transfer to harder cases. Our method achieves superior results across diverse frontier models.
\end{abstract}

\section{Introduction}\label{sec:intro}
Language-model agents are being applied to tasks ranging from everyday information seeking to 
high-stakes decision support. For example, they gather evidence from documents to answer multi-hop questions
\citep{yao2023react,zhou2025mem1} and interpret case-specific findings to answer medical questions
\citep{tang2024medagents,wang2026medagentpro}. Many reasoning tasks require drawing a verifiable conclusion from
case-specific evidence, using domain knowledge and task-specific rules. Examples range from document-based
question answering \citep{zhou2025mem1} and fact verification \citep{yao2023react} to clinical reasoning
\citep{tang2024medagents,wang2026medagentpro} and business reasoning \citep{chen2021finqa}.
Such evidence may be distributed across patient records, multiple documents, or tables and charts.
Agents can be knowledge-rich without being process-aware: recalling facts and rules does not
ensure that they understand how an environment evolves or how actions affect later observations.
Self-evolution offers a way for agents to learn these dynamics from interaction and feedback
and adapt their reasoning to changing conditions
\citep{gao2026selfevolving}. Several high-stake agentic frameworks lack a mechanism for self-evolution: they do not use experience
from earlier cases to improve how they solve new ones. Take medical agents as an example, MedAgents uses specialist discussion
\citep{tang2024medagents}, MDAgents uses complexity-adaptive collaboration \citep{kim2024mdagents},
and MedAgent-Pro combines guideline retrieval with visual tools \citep{wang2026medagentpro}.
These workflows coordinate specialist discussion, retrieval, and tool use within a case, but they do not, by themselves, provide a mechanism for
continual improvement. The broader challenge is to turn experience into reliable supervision
without requiring new human labels \citep{madaan2023selfrefine,zhang2025dgm}. Probing an agent's own decision process
can expose weaknesses in how it uses evidence
and provide feedback for correcting mistakes while preserving sound decisions.

\begin{figure}[t]
  \centering
  \includegraphics[width=\linewidth]{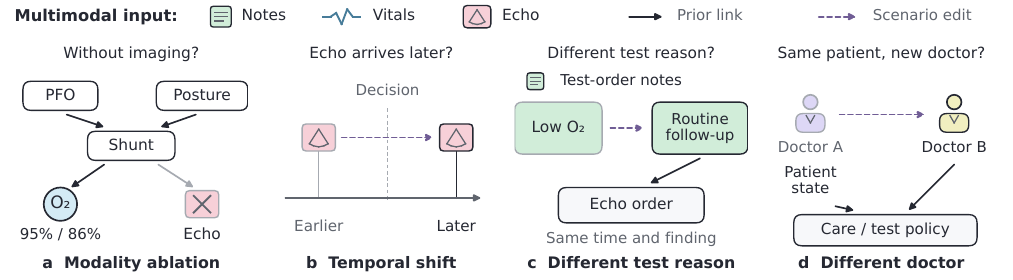}
  \caption{\textbf{A multimodal clinical example with diverse counterfactual context.}
  Using a case from our differential diagnosis evaluation \citep{strickland2023platypnea},
  we illustrate four types of counterfactual context for the agent.
  (a) Removing imaging hides a heart-scan result while retaining oxygen readings
  and prior knowledge of how an opening in the heart (PFO) affects blood flow.
  (b) Shifting the scan result's arrival time tests when its evidence is available
  to the agent.
  (c) Changing the test reason replaces concern about low oxygen with a routine
  checkup in the ordering note, keeping the test time and result fixed.
  (d) Changing the doctor varies the care and test-ordering policy for the same patient.
  These contexts prompt the agent to reason about which evidence is available,
  when and why it was collected, and how clinicians' choices shape it. Verified counterfactuals support self-evolution by providing feedback to train the Proposer and
  reusable context to help the Solver correct errors and preserve sound decisions on later cases.}
  \label{fig:evidence-identifiable}
\end{figure}

Counterfactual (CF) reasoning offers a principled way to probe a decision: vary selected aspects
of the evidence and test how the conclusion depends on them. Recent research has explored
this idea through counterfactual data augmentation, which generates controlled variations of
observed examples to improve generalization and reduce reliance on spurious correlations
\citep{feder2023augmentations}. Other work uses language models to propose health interventions and augment sensor data
\citep{soumma2026counterfactual}. More recently, counterfactuals have been incorporated into
multi-agent clinical reasoning \citep{you2026cfmultiagent} and self-evolution through training
on verified reasoning traces \citep{fan2026iedi}. However, these studies do not directly address
when counterfactual feedback should correct an agent's original decision and when it
should preserve it, without updating the agent's underlying model. A plausible counterfactual that changes an answer does not establish
that the original answer was wrong. Many reasoning tasks provide enough evidence to determine
a definite answer given domain knowledge and task rules. We call these \textbf{evidence-identifiable}
tasks. Figure~\ref{fig:evidence-identifiable} illustrates how counterfactuals test a decision's
dependence on evidence availability, timing, test-ordering reasons, and clinician choices.
Verified alternatives provide feedback beyond the observed case. Retaining this feedback as
context supports self-evolution by helping agents correct recurring reasoning errors while
preserving well-supported decisions.

\begin{figure}[!t]
  \centering
  \includegraphics[width=\linewidth]{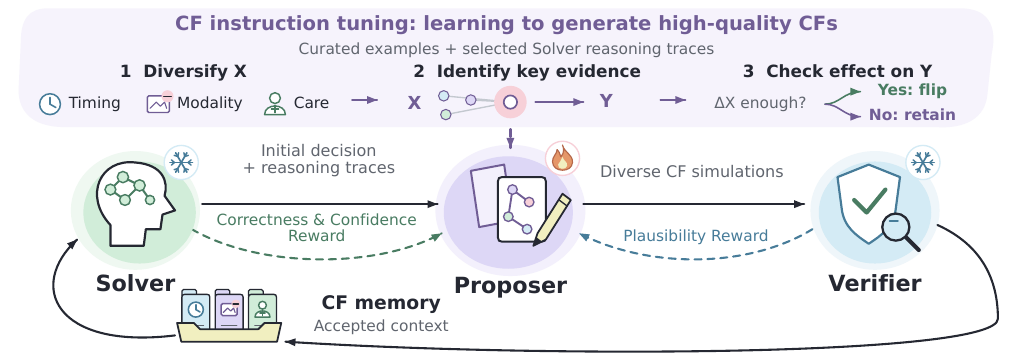}
  \caption{\textbf{Counterfactual self-evolution framework.} The Proposer generates counterfactuals from the Solver's decision and selected reasoning traces. The Verifier checks plausibility, and accepted counterfactuals enter memory as reusable Solver context. The Proposer learns through CF instruction tuning and feedback combining Solver's correctness/confidence with Verifier's plausibility.}
  \label{fig:counterfactual-self-evolution}
\end{figure}

To address these limitations, we propose counterfactual self-evolution (CF-Evolve), a Proposer--Verifier--Solver
framework that improves decision-making through evolving counterfactual context
(Figure~\ref{fig:counterfactual-self-evolution}). The Proposer generates multiple targeted
case variations, the Verifier assesses their plausibility against a knowledge base of clinical
guidelines, medical references, and workflow protocols, and the Solver uses accepted context to
correct errors while reinforcing existing well-supported decisions. All three agents share an LLM backbone; only the Proposer is trained,
while the Solver and Verifier remain frozen, and no stronger teacher is queried during the
evolution loop. We construct a curated counterfactual instruction-tuning dataset spanning diverse
application scenarios across binary classification, multiple-choice reasoning, and regression tasks.
Its structured examples train the Proposer to reason about counterfactuals, apply domain knowledge,
follow a consistent output format, and judge when evidence warrants changing or preserving a decision.
Our instruction targets teach the Proposer to construct counterfactuals, trace their hypothesized
causal mechanisms, and justify whether the original decision should change.
Beyond instruction tuning, we tailor the Proposer's fine-tuning objective to integrate Verifier feedback with
Solver decision quality.
As we progressively inject counterfactual context, we track performance over successive rounds and
find that the framework learns to solve increasingly difficult held-out cases.
We also find that counterfactual instruction tuning can approach the performance of task-specific
Solver fine-tuning with \textbf{substantially fewer training examples}, while leaving the Solver's parameters
unchanged. For example, on MIMIC-IV in-hospital mortality, a Proposer trained on around $3{,}000$
counterfactual examples supplies additional context that brings the frozen Solver's performance close to directly fine-tuning the Solver on $41{,}886$ examples. CF-Evolve thus uses about $1/14$ as many tuning examples. Analyses also show that the generated counterfactual mix
differs across tasks, indicating that useful evidence edits are task-dependent.

\vspace{-1em}
\section{Related Works} \label{sec:related}
Self-evolving agents adapt outputs, memory, code, or model parameters
through feedback. Self-Refine iteratively revises outputs
through self-feedback \citep{madaan2023selfrefine}, while Reflexion retains verbal reflections
to guide later attempts without updating model weights \citep{shinn2023reflexion}. In clinical
reasoning, GSEM organizes experiences into a graph and uses outcome feedback to calibrate
their reliability and joint retrieval \citep{han2026gsem}. Our focus is learning to generate
feedback that tests how decisions depend on evidence.
Other approaches ground improvement in external evaluation. The Darwin G\"odel Machine
modifies agent code and evaluates changes on coding benchmarks \citep{zhang2025dgm}, and
CORAL supports collaborative discovery through shared persistent memory and separate evaluators
\citep{qu2026coral}. Propose, Solve, Verify trains a difficulty-aware problem proposer and
solver through formally verified self-play \citep{wilf2025propose}. Many evidence-based
decisions lack formal correctness checks. Our Verifier instead
checks plausibility and evidence consistency, while Proposer training combines this feedback
with Solver's decision.

Counterfactual methods use controlled changes for explanation, augmentation, and reasoning.
\mbox{FIZLE} prompts LLMs to generate text counterfactuals that probe black-box classifiers
\citep{bhattacharjee2024fizle}. CATO uses causal assumptions to vary spurious attributes for
robust learning \citep{feder2023augmentations}, while SenseCF fine-tunes LLMs for health
intervention design and sensor-data augmentation \citep{soumma2026counterfactual}.
For clinical reasoning, \citet{you2026cfmultiagent} edit findings and measure diagnostic
confidence shifts to guide specialist discussion in a training-free framework. Counterfactuals
also support learning across attempts: EvoCF derives constraints from execution failures,
generates alternative multi-agent plans, and ranks them using retrieved experience
\citep{chi2026evocf}; I-EDI filters reasoning traces through executable counterfactual tests
and structural consistency checks before distilling them into the solver \citep{fan2026iedi}.
CF-Evolve learns a Proposer that edits case evidence and explains whether the original
decision warrants correction or preservation. Training targets distinguish a response to edited
evidence from a justified revision of the original decision. Accepted feedback becomes reusable
context for the frozen Solver. Each memory note records its mechanism, applicability conditions,
and failure boundaries, allowing the Solver to assess whether feedback from an earlier case
is relevant before revising current decision.
\section{Counterfactual Self-Evolving Agents}\label{sec:method}
We now discuss the proposed CF-Evolve method. We begin with a motivating example, then explain counterfactual generation and verification. We next describe CF instruction tuning, feedback-based Proposer learning, and the use of accepted counterfactuals as Solver memory.

\begin{wrapfigure}[19]{r}{0.54\linewidth}
  \vspace{-\intextsep}
  \centering
  \includegraphics[width=\linewidth]{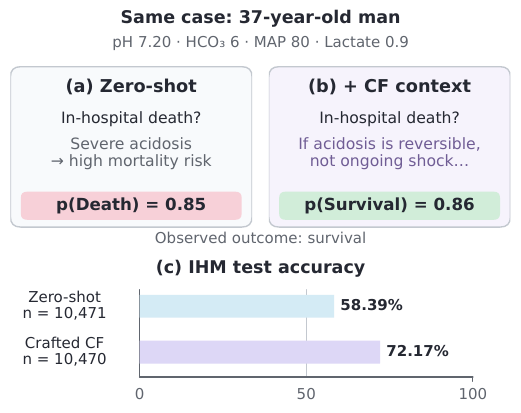}
  \vspace{-2.5em}
  \caption{\small\textbf{CF context supports reconsideration.} (a,b) Crafted CF context corrects an IHM prediction. (c) Test accuracy with zero-shot inference and crafted CF context.}
  \label{fig:counterfactual-motivation}
\end{wrapfigure}
\textbf{Problem Setup and Motivating Example.} 
An evidence-identifiable task asks a question $q$ whose evidence $E$, together with domain knowledge $K$, determines a unique, checkable answer $y$. We write this relation as $f(q,E;K)=y$, where $f$ denotes the task's answer model. We initialize the Solver with a pretrained reasoning model (Qwen2.5-7B-Instruct; \citealp{qwen2025qwen25technicalreport}) that produces a zero-shot prediction and reasoning trace from $q$ and $E$.
Counterfactual context can prompt the Solver to reconsider how evidence supports its decision. Figure~\ref{fig:counterfactual-motivation} illustrates this on in-hospital mortality (IHM) prediction. In the illustrated case, severe acidosis leads the zero-shot Solver to predict death with probability $0.85$. A counterfactual asks whether the acidosis reflects a reversible metabolic process rather than ongoing shock, while preserving the observed findings. The Solver lowers its death probability to $0.14$, correctly predicting survival. The comparison in Figure~\ref{fig:counterfactual-motivation}c shows that crafted CF context improves the Solver's test accuracy. The performance gain motivates learning CF context that helps the Solver reconsider its decisions using the available evidence.

\subsection{Instruction Tuning for Grounded Counterfactual Reasoning}
\label{sec:cf-instruction-tuning}
\textbf{Structural causal formulation.}
We specify the semantics of a CF using an acyclic structural causal model $M=(U,V,F,P_U)$, with background variables $U$, endogenous task variables $V$, mechanisms $F$, and background distribution $P_U$ \citep{pearl2009causality}. Each variable satisfies $V_j=f_j(\mathrm{Pa}_j,U_j)$, where $\mathrm{Pa}_j$ denotes its parents. An action intervention replaces $f_A$ by $A=a'$ while retaining the other mechanisms and the same background state $u$. Consequently,
\begin{equation}
  V_j^{\mathrm{do}(A=a')}(u)=V_j(u)
  \quad\text{for }V_j\notin\{A\}\cup\mathrm{Desc}_M(A).
  \label{eq:cf-local-preservation}
\end{equation}
Here $\mathrm{Desc}_M(A)$ denotes the strict descendants of $A$. The equation states that a variable keeps its original value if the intervention cannot affect it through any causal path. It specifies what an admissible intervention preserves; the correction and preservation analysis below then asks whether the resulting context improves the Solver's answer to the original case. Appendix~\ref{app:cf-scm} formalizes the intervention $A=a'$ and proves that unaffected variables stay fixed.

\textbf{What does a useful counterfactual test?}
Our goal is to design a Proposer that generates CF context to expose reasoning errors while preserving decisions supported by the original evidence. Drawing on causal reasoning \citep{pearl2009causaloverview,pearl2009causality}, we distinguish three tests of evidence, explanation, and action mechanisms in a structural causal model $M$.
\textbf{Observe ($E\!\to\!E'$)} changes which observations are available, or when they become available, to test the Solver's dependence on particular evidence.
\textbf{Explain ($H\!\to\!H'$)} uses abduction to compare causal hypotheses $H$ and $H'$ for the same evidence $E$, testing its interpretation while keeping the observed facts fixed.
\textbf{Intervene ($M\!\to\!M_{\mathrm{do}(A=a')}$)} replaces the mechanism for action $A$ with $A=a'$, leaving other mechanisms unchanged, to test the consequences of a different treatment, support state, or timing.
We aim for balanced coverage of these three types in the Proposer's generated context.

\begin{figure}[t]
  \centering
  \includegraphics[width=\linewidth]{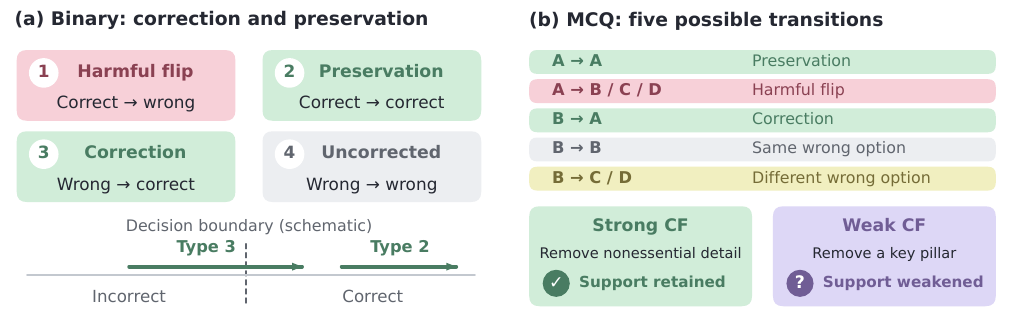}
  \caption{\textbf{Decision scenarios emphasized in CF instruction tuning.} (a) Binary decision transitions favor correction (Type~3) and preservation (Type~2, strengthen the confidence). (b) Multiple-choice questions (MCQ) also allow switches between wrong options; A denotes the correct option. Strong CFs remove nonessential evidence while retaining diagnostic support; weak CFs remove a diagnostic pillar and weaken support without specifying a replacement answer.}
  \label{fig:cf-instruction-tuning}
\end{figure}

\textbf{Correction and preservation.}
For a binary task, let $\hat y$ be the original Solver decision and $\hat y^{+}$ its decision on the original case after reading CF context. Relative to the known training label $y$, there are four outcomes: (1) correct-to-wrong, (2) correct-to-correct, (3) wrong-to-correct, and (4) wrong-to-wrong. Types~2 and~3 are the desired preservation and correction behaviors (Figure~\ref{fig:cf-instruction-tuning}a). Their importance follows from the exact decomposition
\begin{equation}
  \Pr(\hat y^{+}=y)-\Pr(\hat y=y)
  =\Pr(\hat y\ne y,\hat y^{+}=y)
   -\Pr(\hat y=y,\hat y^{+}\ne y).
  \label{eq:correction-preservation}
\end{equation}
The accuracy gain equals the probability of correcting a wrong answer minus the probability of overturning a correct one. A flip-only objective rewards both terms, although they affect accuracy in opposite directions. The early failures in Appendix~\ref{app:cf-failures} illustrate why a changed decision alone is insufficient. The Proposer must therefore assess the support removed, the support that remains, and whether either justifies revising the original decision. These outcome types guide CF instruction-tuning data construction and evaluation.

\textbf{CF instruction-tuning setup.}
CF instruction tuning teaches the Proposer to apply domain knowledge, follow the required output format, and reason about whether changing key evidence warrants revising a decision. We construct supervision from training instances and the target Solver's own predictions, confidence scores, and reasoning traces. Cases with related evidence patterns or reasoning errors form complementary groups: a corrective example identifies an error using the available evidence, a hard preservation example uses counterevidence to rule out an analogous correction, and an anchor preservation example retains a well-supported decision. We add ``\texttt{no\_cf}'' targets when no admissible edit is supported. Labels guide curation but are excluded from Proposer inputs, and factual claims in each target must be supported by the instance's available evidence. Each sample pairs the task, available evidence, and original Solver draft with a structured target: an evidence audit, a local hypothetical edit, its expected effect, an applicability boundary, and optional Solver memory. A separate flip-warrant gate controls whether CF context is used to revise the Solver's decision; corrective, preservation, and abstention targets teach the Proposer to judge when the evidence warrants such a change. Appendix~\ref{app:cf-data} details data construction and checks, Appendix~\ref{app:cf-target} explains the target fields, and Appendix~\ref{app:cf-training} gives the training objective and settings.

We next analyze the behavior these instruction targets aim to teach. The two propositions below separate net accuracy gain from harmful reversals and characterize when CF context is worth using.

\textbf{Proposition 1 (Causal benefit of CF context).}
For fixed Solver and CF-generation procedures, let $R_t=\mathbf{1}\{\hat y(t)=y\}$ denote Solver correctness without ($t=0$) or with ($t=1$) CF context, holding the original case fixed. Correction $b=\Pr(R_1=1,R_0=0)$ is a probability of necessity and sufficiency \citep{tian2000probabilities}; harm is $h=\Pr(R_1=0,R_0=1)$. For marginal accuracies $p_t=\Pr(R_t=1)$, the sharp bounds are
\begin{equation}
  \max(0,p_1-p_0)\le b\le\min(p_1,1-p_0),
  \qquad h=b-(p_1-p_0).
  \label{eq:cf-benefit-bounds}
\end{equation}
Even with higher accuracy, some originally correct answers may be harmed; paired correction and harm rates reveal what the net gain conceals.

\textbf{Proposition 2 (When to propose).}
Let $X$ collect the available evidence, original Solver draft, and candidate CF, excluding the reference answer, and let $b(X)$ and $h(X)$ be the conditional correction and harm probabilities. Suppose abstention retains the original decision, correction earns utility $\kappa_+>0$, harm costs $\kappa_->0$, and unchanged correctness has zero incremental utility. Then the inject-or-abstain rule maximizing expected incremental utility is
\begin{equation}
  g^*(X)=\mathbf{1}\{\kappa_+b(X)>\kappa_-h(X)\},
  \label{eq:cf-selective-rule}
\end{equation}
with ties resolved by abstention. The rule \textbf{uses CF context only when the expected benefit of correction exceeds the expected cost of harm}; with equal weights, it selects positive expected accuracy gain. It describes the ideal behavior our instruction targets and flip warrant seek to approximate, without assuming that the implemented gate estimates calibrated response probabilities. Appendix~\ref{app:cf-causal-analysis} gives the assumptions and proofs of both propositions.

\subsection{Self-Evolution via Feedback from Multi-Agents}
\label{sec:memory-evolution}
Instruction tuning does not directly test whether new proposals are plausible or help the Solver answer correctly. Verifier assessments and Solver feedback computed from training labels further train the Proposer, while the Solver revises its answers using the resulting CF context. In later memory-evolution rounds, no agent is retrained; only the bank of accepted CFs grows.

\textbf{Verifier feedback.}
The Verifier checks proposed edits against retrieved domain knowledge. As an example, our clinical implementation uses Chroma with nine topic-specific PubMed Central collections, DailyMed drug labels, Surviving Sepsis Campaign guidelines, and ACR imaging criteria, covering major ICU physiological systems. Topic routing and semantic retrieval supply evidence to a judge that returns PASS, FLAG, or REJECT verdicts. Appendix~\ref{app:collective-feedback} details retrieval and scoring.

\textbf{Solver feedback and memory.}
The memory bank organizes factual cases, verified CFs, and knowledge gaps with task tags and provenance. Reusable notes retain a mechanism, applicability conditions, and failure boundaries. The frozen Solver reads the original case and draft alongside the current CF note and separately marked retrieved experience. Its revised answer distribution $p^c$ is compared with the original $p^0$. On training cases, the reference answer $y$ supplies both a correctness signal $r_C$ and an uncertainty signal $r_U$: the clipped reduction in true-answer log loss. Thus confidence is rewarded when it moves toward the correct answer. To prevent context explosion, the memory bank imposes a fixed capacity on the CF context. The feedback objective therefore optimizes the selection, ordering, and combination of CF contexts within this budget, favoring context that is relevant to the current case and useful for improving the Solver's decision.

\textbf{Feedback-based Proposer optimization.}
We formulate the collective-feedback objective as a correctness-veto format. For training pairs $(x,y)\sim\mathcal D_{\mathrm{train}}$ and CFs $c\sim\pi_\theta(\cdot\mid x)$, let $g=\mathbf 1\{\hat y^c=y\}$ indicate a correct revised answer and $s=\lambda_U r_U+\lambda_V r_V+r_F$ combine auxiliary feedback, with $\lambda_C>0$ and $\lambda_U,\lambda_V\ge0$. Then the objective is:
\begin{equation}
  \max_\theta\quad J(\theta)=\mathbb E\!\left[\lambda_C r_C+g\,s+(1-g)\min(s,0)\right],
\label{eq:collective-feedback-objective}
\end{equation}
where $r_C$ is positive for correction and preservation and negative for harmful reversals and unresolved errors. The uncertainty $r_U=\mathrm{clip}(\log p^c(y)-\log p^0(y),-2,2)$ rewards increased probability of the true answer and penalizes decreases, with its magnitude capped at two. Verifier support is scored by $r_V$, and $r_F$ combines the implemented direction/schema bonuses and invalid-output penalties. When the revised answer is wrong, the combined auxiliary contribution is capped at zero, so the total reward remains negative. Labels are only used in reward computation. 

\textbf{Memory evolution}
lets the frozen Solver reuse validated experience on new cases without retraining. Starting from $\mathcal M_0=\emptyset$, each round applies the trained agents and CF admission gate to disjoint new evolution cases, storing accepted notes $\mathcal A_r$ as $\mathcal M_r=\mathcal M_{r-1}\cup\mathcal A_r$ for $r=1,\ldots,R$. The Solver receives selected CF context within a fixed budget. Context evolution has several dimensions: the number of notes within that budget, diversity of CF types, coverage of source subgroups, and the quality and relevance of the selected context. Feedback-based Proposer optimization targets context quality by rewarding its contribution to the Solver's decision. Fixed evaluation sets track each difficulty levels against its memory-free reference. The following bound assesses whether these changes yield more corrections than harmful reversals, accounting for sampling uncertainty.

\textbf{Proposition 3 (Reliable memory evolution).}
For $R$ prespecified rounds and $S$ level, let $\Delta_{r,s}$ be the population accuracy gain from $\mathcal M_{r-1}$ to $\mathcal M_r$ under a fixed Solver context budget, estimated by the mean paired gain $\widehat\Delta_{r,s}$ on $n_s$ independent cases. If the entire memory trajectory, including its context-selection and ordering rules, is independent of the evaluation sets, then for $\delta\in(0,1)$,
\begin{equation}
  L_{r,s}=\widehat\Delta_{r,s}-\sqrt{\frac{2\log(RS/\delta)}{n_s}},
  \qquad
  \Pr\!\left(\Delta_{r,s}\ge L_{r,s}\ \forall r,s\right)\ge1-\delta.
  \label{eq:memory-evolution-lcb}
\end{equation}
Thus $\min_s L_{r,s}>0$ certifies improvement across difficulty level. Appendix~\ref{app:memory-evolution} gives the assumptions and proof using Hoeffding's inequality \citep{hoeffding1963probability}. 
Specifically, the bound subtracts a margin for sampling uncertainty from the observed gain at each difficulty level. If every remainder is positive, the context update improves all levels with confidence at least $1-\delta$. 

\section{Experiments}\label{sec:exp}

We evaluate CF-Evolve across healthcare, business, and fact verification domains. We ask the following research questions (RQ): (\textbf{RQ1}) How does counterfactual instruction tuning for Proposer improve CF quality? (\textbf{RQ2}) How robust is our framework across base models and component ablations, and how does it compare with strong baselines? (\textbf{RQ3}) Can CF-driven self-evolution progressively address harder problems across diverse situations?

\textbf{Datasets and Baselines.} We evaluate on the MIMIC-IV ecosystem \citep{johnson2023mimiciv} for multimodal prediction of in-hospital mortality (IHM) and length of stay (LOS), DiagnosisArena \citep{zhu2025diagnosisarena} for differential diagnosis, the MMMU Business section \citep{yue2024mmmu}, with an additional
MMMU/MMMU-Pro business evaluation for memory transfer, and HoVer \citep{jiang2020hover} for multi-hop fact verification. These tasks span binary classification, MCQ, and regression. We identify three baseline groups: (i) neither CF nor self-evolution, including zero-shot and task-fine-tuned base models, plus FuseMoE \citep{han2024fusemoe} and MILM \citep{chung2026milm} for clinical prediction; (ii) CF without self-evolution, using a static retrieval model to supply relevant CFs from a fixed bank as Solver context \citep{zhou2023contextfaithful}; and (iii) self-evolution without CF, such as GSEM \citep{han2026gsem}, which updates an experience memory through feedback.

\textbf{Experiment Setup.}
We evaluate Qwen2.5-7B \citep{qwen2025qwen25technicalreport}, Llama3-8B
\citep{meta2024llama3}, and MedGemma-4B \citep{sellergren2025medgemma}.
Task- and model-specific LoRA Proposers condition on the evidence and the target
Solver's initial prediction and reasoning, while the Solver and Verifier remain
frozen. Validation data select MCQ checkpoints and CF-acceptance thresholds.
We report ranking and classification metrics for IHM, absolute error for LOS,
and accuracy for MCQ and fact verification; paired corrected and harmed counts
measure decision revisions.

\begin{figure*}[t]
\centering
\includegraphics[width=\linewidth]{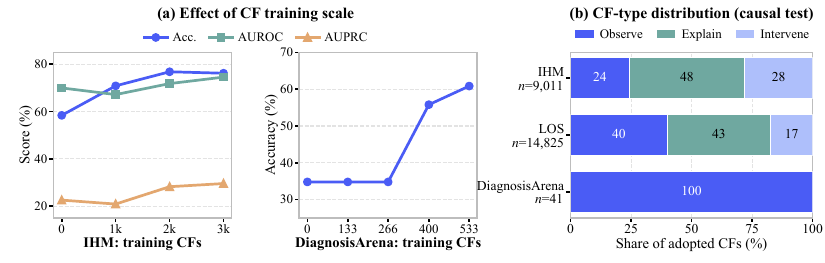}
\vspace{-2mm}
\caption{\textbf{CF instruction tuning produces high-quality CF context.} (a) We instruction-tune a Qwen2.5-7B Proposer on increasing numbers of training CFs to guide a frozen Solver. (b) We categorize the adopted CF context by the causal test it performs mentioned in Section~\ref{sec:cf-instruction-tuning}: \emph{Observe} changes the available evidence or its timing; \emph{Explain} changes the causal hypothesis while holding evidence fixed; and \emph{Intervene} changes an action. Scores are shown on a 0--100 scale.\looseness=-1}
\label{tab:rq1_cfit_main}
\vspace{-4mm}
\end{figure*}

\subsection{CF Instruction Tuning Shapes Decision Quality (RQ1)}

Figure~\ref{tab:rq1_cfit_main} shows that high-quality CF context depends on both supervision
and the available causal tests. With limited supervision, higher IHM accuracy
accompanies lower AUROC and AUPRC, meaning that predicted risks distinguish deaths
from survivors less well. The 1k set is randomly sampled, whereas the 2k--3k sets
use curated examples, so gains may reflect better examples as well as more data.
On DiagnosisArena, the gain from 400 to 533
training records comes entirely from fewer harmful revisions: both runs correct
36 errors, but the latter preserves every initially correct answer
(Appendix~\ref{app:diag-cf-it}). This supports our emphasis on teaching the Proposer
when to preserve a decision alongside how to correct it. The CF-type distribution
also reflects the task and its edit interface: clinical outputs can test alternative
explanations and actions, while DiagnosisArena uses executable edits to observed
evidence. These tests probe the assumptions supporting the original decision while
keeping hypothetical changes distinct from observed facts. Our framework accommodates
these different tests within the same correction--preservation objective, allowing
the Proposer to improve the guidance supplied to a frozen Solver.\looseness=-1 

\subsection{CF-Evolve Generalizes across Tasks and Base Models (RQ1-2)}
\begin{figure*}[t]
\centering
\includegraphics[width=\linewidth]{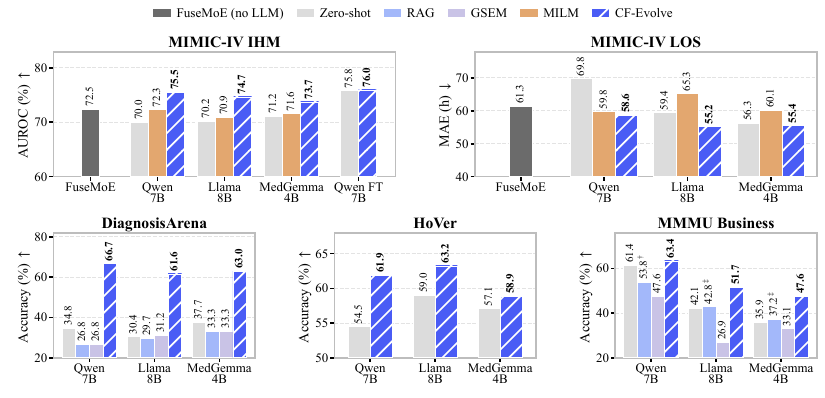}
\vspace{-2mm}
\caption{\textbf{Comprehensive evaluations of CF-Evolve.} We instruction-tune diverse base models to generate CF context that guides a frozen Solver, followed by memory evolution to optimize the involved context. Results show that adding high-quality CF context consistently improves performance across tasks. Task-specific baselines are shown only where applicable.}
\label{tab:rq2_robustness}
\vspace{-4mm}
\end{figure*}

Figure~\ref{tab:rq2_robustness} evaluates whether CF-Evolve remains useful when the
task and base model change. We compare each configuration against its corresponding
baselines. MCQ configurations use separately trained Proposers and validation-selected
acceptance policies, allowing the same framework to accommodate different Solver
behaviors.

CF-Evolve improves the reported metric in every task--model setting shown and
outperforms the competing baselines where they are evaluated. Gains span clinical
prediction, diagnostic reasoning, and non-medical reasoning, supporting the
framework's applicability beyond a single domain. The Proposer tests whether changing
a finding or an assumption would change the answer, then explains what this reveals
about the Solver's original reasoning. The Solver uses that explanation to reconsider
its answer without updating its weights. This feedback yields more consistent MCQ
gains than the RAG and GSEM baselines. Even for the Qwen Solver fine-tuned on 41,886
examples, a Proposer trained on 3,000 CF examples slightly improves IHM AUROC. This
supports CF instruction tuning as a sample-efficient complement to direct Solver
training. Full metric
vectors, paired correction--harm counts, and task-specific configurations can be found in
Appendix~\ref{app:rq2-extended}.

\subsection{CF-Evolve Progressively Solves Harder Problems (RQ3)}

\begin{figure}[t]
  \centering
  \includegraphics[width=\linewidth]{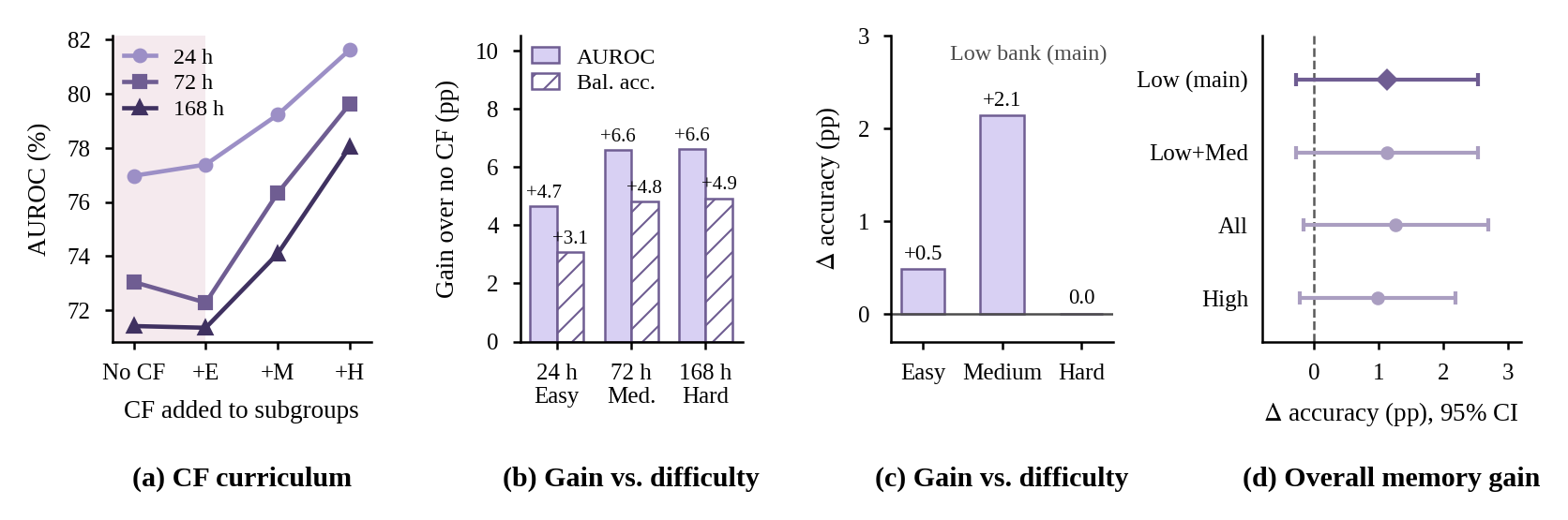}
  \caption{\textbf{Evaluate CF-Evolve on harder problems.} (a, b) CF-Evolve accumulates memory across patient-difficulty subgroups for clinical mortality prediction at increasing prediction horizons. (c, d) CF-Evolve accumulates memory across CF banks for MMMU/MMMU-Pro business questions of increasing difficulty. Across both domains, richer context yields limited gains on easier problems but increasingly benefits harder ones. Easy decisions already have sufficient evidence, whereas difficult cases benefit from counterfactual experience that supplies missing or ambiguous reasoning paths.}
  \label{fig:rq3}
\end{figure}

We further evaluate whether expanding CF context, and thus exposure to alternative
evidence--decision and action--outcome relations, helps a frozen Solver solve harder problems.

\textbf{Harder patient cases.} From the same first-24-hour record, we predict
mortality at increasingly difficult 24-, 72-, and 168-hour horizons
(Appendix~\ref{app:rq3-curriculum}). Patient-specific CF raises AUROC by 4.65,
6.57, and 6.62 points, respectively; balanced accuracy follows the same pattern
(Figure~\ref{fig:rq3}b). The hardest task's AUROC with CF exceeds the easiest task's baseline.
Within each task, we progressively supply CF to confidence-defined Easy, Medium,
and Hard patient groups. The Medium and Hard rounds account for at least 91\% of
the final AUROC gain, whereas Easy-only CF changes little
(Figure~\ref{fig:rq3}a). This pattern suggests that CF is most useful when the
Solver needs help resolving uncertain evidence, supporting targeted context
expansion for harder decisions.

\textbf{Harder business reasoning.} We evaluate CF memory on unseen MMMU/MMMU-Pro
business questions with predefined Easy, Medium, and Hard difficulty levels.
We group source cases by initial Solver confidence to form four banks: Low,
Low+Medium, All, and High-only. All four yield positive accuracy gains, although
their confidence intervals include zero (Figure~\ref{fig:rq3}d). The
validation-selected Low bank raises accuracy from 56.48\% to 57.61\%, helping
Medium questions most
(Figure~\ref{fig:rq3}c); larger banks yield small gains on Hard questions
(Appendix~\ref{app:rq3-memory}). The memory audit suggests a reason for this
uneven transfer: most stored CFs preserve the source answer and contain
source-specific numbers. Such examples can reinforce familiar reasoning, but may
offer limited guidance when a new question requires different decision dependencies.
Together, these results support expanding relevant CF coverage for difficult cases
and suggest that transfer depends on capturing reusable reasoning, beyond increasing
the number of memories.
\section{Conclusion and Future Work}\label{sec:conclusion}
We introduced counterfactual self-evolution, combining grounded counterfactual instruction tuning
with memory-based self-evolution for a frozen Solver. The Proposer learns to test how decisions
depend on evidence, helping the Solver correct reasoning errors and retain answers supported by the
original evidence. Different CF categories distinguish changes to observations, causal hypotheses, and
actions. The Proposer is refined by collecting feedback from Solver's correctness and uncertainty combined with Verifier's feedback.
Accepted counterfactual accumulates as reusable context, so the Solver
adapts without further fine-tuning. Our analysis clarifies intervention semantics and gives conditions
under which correction outweighs harm and memory updates improve performance. The results
suggest that broader CF coverage helps difficult clinical decisions, while transfer across cases
depends on whether memory captures reusable reasoning. Future
work could estimate correction and harm probabilities to guide counterfactual selection and use
uncertainty bounds to control memory updates. A second direction is to learn structural causal models that constrain
edits and actively select informative counterfactuals to resolve uncertain mechanisms. They
could improve how agents acquire and transfer experience as task environments
change.

\bibliography{reference}

@article{pearl2009causaloverview,
  title={Causal Inference in Statistics: An Overview},
  author={Pearl, Judea},
  journal={Statistics Surveys},
  volume={3},
  pages={96--146},
  year={2009},
  doi={10.1214/09-SS057}
}

@inproceedings{wilf2025propose,
  title={Propose, Solve, Verify: Self-Play Through Formal Verification},
  author={Wilf, Alex and Aggarwal, Pranjal and Parno, Bryan and Fried, Daniel and Morency, Louis-Philippe and Liang, Paul Pu and Welleck, Sean},
  booktitle={Forty-third International Conference on Machine Learning},
  year={2026},
  url={https://openreview.net/forum?id=Vuaq1qXCki}
}

@article{qu2026coral,
  title={{CORAL}: Towards Autonomous Multi-Agent Evolution for Open-Ended Discovery},
  author={Qu, Ao and Zheng, Han and Zhou, Zijian and Yan, Yihao and Tang, Yihong and Ong, Shao Yong and Hong, Fenglu and Zhou, Kaichen and Jiang, Chonghe and Kong, Minwei and Zhu, Jiacheng and Jiang, Xuan and Li, Sirui and Wu, Cathy and Low, Bryan Kian Hsiang and Zhao, Jinhua and Liang, Paul Pu},
  journal={arXiv preprint arXiv:2604.01658},
  year={2026},
  url={https://arxiv.org/abs/2604.01658}
}

@article{soumma2026counterfactual,
  title={Counterfactual Modeling with Fine-Tuned LLMs for Health Intervention Design and Sensor Data Augmentation},
  author={Soumma, Shovito Barua and Arefeen, Asiful and Carpenter, Stephanie M and Hingle, Melanie and Ghasemzadeh, Hassan},
  journal={IEEE Open Journal of Engineering in Medicine and Biology},
  volume={7},
  pages={232--240},
  year={2026},
  doi={10.1109/OJEMB.2026.3697994},
  url={https://doi.org/10.1109/OJEMB.2026.3697994}
}

@inproceedings{tang2024medagents,
  title={{M}ed{A}gents: Large Language Models as Collaborators for Zero-shot Medical Reasoning},
  author={Tang, Xiangru and Zou, Anni and Zhang, Zhuosheng and Li, Ziming and Zhao, Yilun and Zhang, Xingyao and Cohan, Arman and Gerstein, Mark},
  booktitle={Findings of the Association for Computational Linguistics: ACL 2024},
  year={2024},
  pages="599--621",
  doi="10.18653/v1/2024.findings-acl.33",
  url="https://aclanthology.org/2024.findings-acl.33/"
}

@inproceedings{kim2024mdagents,
  title={{MDAgents}: An Adaptive Collaboration of {LLM}s for Medical Decision-Making},
  author={Kim, Yubin and Park, Chanwoo and Jeong, Hyewon and Chan, Yik Siu and Xu, Xuhai and McDuff, Daniel and Lee, Hyeonhoon and Ghassemi, Marzyeh and Breazeal, Cynthia and Park, Hae Won},
  booktitle={Advances in Neural Information Processing Systems (NeurIPS)},
  year={2024},
  pages={79410--79452},
  volume={37},
  doi={10.52202/079017-2522},
  url={https://proceedings.neurips.cc/paper_files/paper/2024/file/90d1fc07f46e31387978b88e7e057a31-Paper-Conference.pdf}
}

@inproceedings{wang2026medagentpro,
  title={{MedAgent-Pro}: Towards Evidence-based Multi-modal Medical Diagnosis via Reasoning Agentic Workflow},
  author={Wang, Ziyue and Wu, Junde and Cai, Linghan and Low, Chang Han and Yang, Xihong and Li, Qiaxuan and Jin, Yueming},
  booktitle={International Conference on Learning Representations},
  pages={114360--114389},
  year={2026},
  url={https://proceedings.iclr.cc/paper_files/paper/2026/file/ba1c5356d9164bb64c446a4b690226b0-Paper-Conference.pdf}
}

@inproceedings{madaan2023selfrefine,
  title={Self-Refine: Iterative Refinement with Self-Feedback},
  author={Madaan, Aman and Tandon, Niket and Gupta, Prakhar and Hallinan, Skyler and Gao, Luyu and Wiegreffe, Sarah and Alon, Uri and Dziri, Nouha and Prabhumoye, Shrimai and Yang, Yiming and Gupta, Shashank and Majumder, Bodhisattwa Prasad and Hermann, Katherine and Welleck, Sean and Yazdanbakhsh, Amir and Clark, Peter},
  booktitle={Advances in Neural Information Processing Systems (NeurIPS)},
  year={2023},
  pages={46534--46594},
  volume={36},
  doi={10.52202/075280-2019},
  url={https://proceedings.neurips.cc/paper_files/paper/2023/file/91edff07232fb1b55a505a9e9f6c0ff3-Paper-Conference.pdf}
}

@inproceedings{shinn2023reflexion,
  title={Reflexion: Language Agents with Verbal Reinforcement Learning},
  author={Shinn, Noah and Cassano, Federico and Gopinath, Ashwin and Narasimhan, Karthik and Yao, Shunyu},
  booktitle={Advances in Neural Information Processing Systems (NeurIPS)},
  year={2023},
  pages={8634--8652},
  volume={36},
  doi={10.52202/075280-0377},
  url={https://proceedings.neurips.cc/paper_files/paper/2023/file/1b44b878bb782e6954cd888628510e90-Paper-Conference.pdf}
}

@inproceedings{zhang2025dgm,
  title={Darwin {G\"odel} Machine: Open-Ended Evolution of Self-Improving Agents},
  author={Zhang, Jenny and Hu, Shengran and Lu, Cong and Lange, Robert and Clune, Jeff},
  booktitle={International Conference on Learning Representations (ICLR)},
  year={2026},
  url={https://openreview.net/forum?id=pUpzQZTvGY}
}

@inproceedings{zhou2025mem1,
  title={{MEM1}: Learning to Synergize Memory and Reasoning for Efficient Long-Horizon Agents},
  author={Zhou, Zijian and Qu, Ao and Wu, Zhaoxuan and Kim, Sunghwan and Prakash, Alok and Rus, Daniela and Zhao, Jinhua and Low, Bryan Kian Hsiang and Liang, Paul Pu},
  booktitle={The Fourteenth International Conference on Learning Representations},
  year={2026},
  url={https://openreview.net/forum?id=XY8AaxDSLb}
}

@inproceedings{feder2023augmentations,
  title={Data Augmentations for Improved (Large) Language Model Generalization},
  author={Feder, Amir and Wald, Yoav and Shi, Claudia and Saria, Suchi and Blei, David},
  booktitle={Advances in Neural Information Processing Systems (NeurIPS)},
  volume={36},
  pages={70638--70653},
  year={2023},
  doi={10.52202/075280-3096},
  url={https://proceedings.neurips.cc/paper_files/paper/2023/file/df88b275bef31ac96c85f0c4013734fc-Paper-Conference.pdf}
}

@inproceedings{bhattacharjee2024fizle,
  title={Zero-shot {LLM}-guided Counterfactual Generation: A Case Study on {NLP} Model Evaluation},
  author={Bhattacharjee, Amrita and Moraffah, Raha and Garland, Joshua and Liu, Huan},
  booktitle={2024 IEEE International Conference on Big Data (BigData)},
  pages={1243--1248},
  year={2024},
  doi={10.1109/BigData62323.2024.10825537},
  url={https://doi.org/10.1109/BigData62323.2024.10825537}
}

@article{you2026cfmultiagent,
  title={Improving Clinical Diagnosis with Counterfactual Multi-Agent Reasoning},
  author={You, Zhiwen and Chen, Xi and Vashishtha, Aniket and Du, Simo and Erion-Barner, Gabriel and Mei, Hongyuan and Peng, Hao and Guo, Yue},
  journal={arXiv preprint arXiv:2603.27820},
  year={2026},
  url={https://arxiv.org/abs/2603.27820}
}

@article{akbarian2024quadratic,
  title={Quadratic Gating Mixture of Experts: Statistical Insights into Self-Attention},
  author={Akbarian, Pedram and Nguyen, Huy and Han, Xing and Ho, Nhat},
  journal={arXiv preprint arXiv:2410.11222},
  year={2024},
  url={https://arxiv.org/abs/2410.11222}
}

@article{nguyen2024hierarchical,
  title={On Expert Estimation in Hierarchical Mixture of Experts: Beyond Softmax Gating Functions},
  author={Nguyen, Huy and Han, Xing and Harris, Carl William and Saria, Suchi and Ho, Nhat},
  journal={arXiv preprint arXiv:2410.02935},
  year={2024},
  url={https://arxiv.org/abs/2410.02935v2}
}

@inproceedings{han2026merge,
  title={Massively Multimodal Foundation Models: A Framework for Capturing Interactions with Specialized Mixture-of-Experts},
  author={Han, Xing and Chung, Hsing-Huan and Ghosh, Joydeep and Liang, Paul Pu and Saria, Suchi},
  booktitle={The Fourteenth International Conference on Learning Representations},
  year={2026},
  url={https://openreview.net/forum?id=qF9WJxvHX8}
}

@inproceedings{han2024fusemoe,
  title={{FuseMoE}: Mixture-of-Experts Transformers for Fleximodal Fusion},
  author={Han, Xing and Nguyen, Huy and Harris, Carl and Ho, Nhat and Saria, Suchi},
  booktitle={Advances in Neural Information Processing Systems (NeurIPS)},
  year={2024},
  pages={67850--67900},
  volume={37},
  doi={10.52202/079017-2167},
  url={https://proceedings.neurips.cc/paper_files/paper/2024/file/7d62a85ebfed2f680eb5544beae93191-Paper-Conference.pdf}
}

@article{chung2026milm,
  title={{MILM}: Large Language Models for Multimodal Irregular Time Series with Informative Sampling},
  author={Chung, Hsing-Huan and Li, Shijun and Wald, Yoav and Han, Xing and Saria, Suchi and Ghosh, Joydeep},
  journal={arXiv preprint arXiv:2605.13711},
  year={2026},
  url={https://arxiv.org/abs/2605.13711}
}

@inproceedings{zhu2025diagnosisarena,
  title={{D}iagnosis{A}rena: Benchmarking Diagnostic Reasoning for Large Language Models},
  author={Zhu, Yakun and Huang, Zhongzhen and Mu, Linjie and Huang, Yutong and Nie, Wei and Liu, Jiaji and Zhang, Shaoting and Liu, Pengfei and Zhang, Xiaofan},
  booktitle={Findings of the Association for Computational Linguistics: ACL 2026},
  pages={3074--3098},
  publisher={Association for Computational Linguistics},
  year={2026},
  doi={10.18653/v1/2026.findings-acl.151},
  url={https://aclanthology.org/2026.findings-acl.151/}
}

@inproceedings{chen2021finqa,
  title={{F}in{QA}: A Dataset of Numerical Reasoning over Financial Data},
  author={Chen, Zhiyu and Chen, Wenhu and Smiley, Charese and Shah, Sameena and Borova, Iana and Langdon, Dylan and Moussa, Reema and Beane, Matt and Huang, Ting-Hao and Routledge, Bryan and Wang, William Yang},
  booktitle={Proceedings of the 2021 Conference on Empirical Methods in Natural Language Processing},
  pages={3697--3711},
  year={2021},
  doi={10.18653/v1/2021.emnlp-main.300},
  url={https://aclanthology.org/2021.emnlp-main.300/}
}

@inproceedings{yao2023react,
  title={{ReAct}: Synergizing Reasoning and Acting in Language Models},
  author={Yao, Shunyu and Zhao, Jeffrey and Yu, Dian and Du, Nan and Shafran, Izhak and Narasimhan, Karthik and Cao, Yuan},
  booktitle={International Conference on Learning Representations (ICLR)},
  year={2023}
}

@article{gao2026selfevolving,
  title={A Survey of Self-Evolving Agents: What, When, How, and Where to Evolve on the Path to Artificial Super Intelligence},
  author={Gao, Huan-ang and Geng, Jiayi and Hua, Wenyue and Hu, Mengkang and Juan, Xinzhe and Liu, Hongzhang and Liu, Shilong and Qiu, Jiahao and Qi, Xuan and Ren, Qihan and Wu, Yiran and Wang, Hongru and Xiao, Han and Zhou, Yuhang and Zhang, Shaokun and Zhang, Jiayi and Xiang, Jinyu and Fang, Yixiong and Zhao, Qiwen and Liu, Dongrui and Qian, Cheng and Wang, Zhenhailong and Hu, Minda and Wang, Huazheng and Wu, Qingyun and Ji, Heng and Wang, Mengdi},
  journal={Transactions on Machine Learning Research},
  year={2026},
  url={https://openreview.net/forum?id=CTr3bovS5F}
}

@inproceedings{fan2026iedi,
  title={{I-EDI}: Robust Self-Evolution Agents via Verifiable Counterfactual Simulation},
  author={Fan, Runze and Li, Yong},
  booktitle={Proceedings of the Thirty-Fifth International Joint Conference on Artificial Intelligence (IJCAI)},
  pages={90--98},
  year={2026},
  doi={10.24963/ijcai.2026/11},
  url={https://doi.org/10.24963/ijcai.2026/11}
}

@article{strickland2023platypnea,
  title={Platypnea-Orthodeoxia Syndrome: A Rare Cause of Hypoxia},
  author={Strickland, Emily and Kerosky, Zachary and Krivda, Joseph and Arrey-Mbi, Takor},
  journal={Annals of Internal Medicine: Clinical Cases},
  volume={2},
  number={2},
  pages={e221124},
  year={2023},
  doi={10.7326/aimcc.2022.1124},
  url={https://doi.org/10.7326/aimcc.2022.1124}
}

@article{han2026gsem,
  title={{GSEM}: Graph-based Self-Evolving Memory for Experience Augmented Clinical Reasoning},
  author={Han, Xiao and Fan, Yuzheng and Zhao, Sendong and Wang, Haochun and Qin, Bing},
  journal={arXiv preprint arXiv:2603.22096},
  year={2026},
  url={https://arxiv.org/abs/2603.22096}
}

@article{qwen2025qwen25technicalreport,
  title={{Qwen2.5} Technical Report},
  author={Yang, An and Yang, Baosong and Zhang, Beichen and Hui, Binyuan and Zheng, Bo and Yu, Bowen and Li, Chengyuan and Liu, Dayiheng and Huang, Fei and Wei, Haoran and Lin, Huan and Yang, Jian and Tu, Jianhong and Zhang, Jianwei and Yang, Jianxin and Yang, Jiaxi and Zhou, Jingren and Lin, Junyang and Dang, Kai and Lu, Keming and Bao, Keqin and Yang, Kexin and Yu, Le and Li, Mei and Xue, Mingfeng and Zhang, Pei and Zhu, Qin and Men, Rui and Lin, Runji and Li, Tianhao and Tang, Tianyi and Xia, Tingyu and Ren, Xingzhang and Ren, Xuancheng and Fan, Yang and Su, Yang and Zhang, Yichang and Wan, Yu and Liu, Yuqiong and Cui, Zeyu and Zhang, Zhenru and Qiu, Zihan},
  journal={arXiv preprint arXiv:2412.15115},
  year={2025},
  doi={10.48550/arXiv.2412.15115},
  url={https://arxiv.org/abs/2412.15115}
}

@book{pearl2009causality,
  title={Causality: Models, Reasoning, and Inference},
  author={Pearl, Judea},
  edition={2},
  publisher={Cambridge University Press},
  year={2009},
  doi={10.1017/CBO9780511803161},
  isbn={9780521895606}
}

@article{tian2000probabilities,
  title={Probabilities of Causation: Bounds and Identification},
  author={Tian, Jin and Pearl, Judea},
  journal={Annals of Mathematics and Artificial Intelligence},
  volume={28},
  number={1--4},
  pages={287--313},
  year={2000},
  doi={10.1023/A:1018912507879}
}

@inproceedings{thomas2015highconfidence,
  title={High Confidence Policy Improvement},
  author={Thomas, Philip and Theocharous, Georgios and Ghavamzadeh, Mohammad},
  booktitle={Proceedings of the 32nd International Conference on Machine Learning},
  series={Proceedings of Machine Learning Research},
  volume={37},
  pages={2380--2388},
  year={2015},
  publisher={PMLR},
  url={https://proceedings.mlr.press/v37/thomas15.html}
}

@article{hoeffding1963probability,
  title={Probability Inequalities for Sums of Bounded Random Variables},
  author={Hoeffding, Wassily},
  journal={Journal of the American Statistical Association},
  volume={58},
  number={301},
  pages={13--30},
  year={1963},
  doi={10.1080/01621459.1963.10500830}
}

@misc{meta2024llama3,
  title={Meta {Llama 3} Model Card},
  author={{Meta AI}},
  year={2024},
  url={https://huggingface.co/meta-llama/Meta-Llama-3-8B-Instruct}
}

@article{sellergren2025medgemma,
  title={{MedGemma} Technical Report},
  author={Sellergren, Andrew and Kazemzadeh, Sahar and Jaroensri, Tiam and others},
  journal={arXiv preprint arXiv:2507.05201},
  year={2026},
  doi={10.48550/arXiv.2507.05201},
  url={https://arxiv.org/abs/2507.05201}
}

@article{johnson2023mimiciv,
  title={{MIMIC-IV}, a freely accessible electronic health record dataset},
  author={Johnson, Alistair E. W. and Bulgarelli, Lucas and Shen, Lu and Gayles, Alvin and Shammout, Ayad and Horng, Steven and Pollard, Tom J. and Hao, Sicheng and Moody, Benjamin and Gow, Brian and Lehman, Li-wei H. and Celi, Leo A. and Mark, Roger G.},
  journal={Scientific Data},
  volume={10},
  pages={1},
  year={2023},
  doi={10.1038/s41597-022-01899-x},
  url={https://doi.org/10.1038/s41597-022-01899-x}
}

@inproceedings{yue2024mmmu,
  title={{MMMU}: A Massive Multi-discipline Multimodal Understanding and Reasoning Benchmark for Expert {AGI}},
  author={Yue, Xiang and Ni, Yuansheng and Zhang, Kai and Zheng, Tianyu and Liu, Ruoqi and Zhang, Ge and Stevens, Samuel and Jiang, Dongfu and Ren, Weiming and Sun, Yuxuan and Wei, Cong and Yu, Botao and Yuan, Ruibin and Sun, Renliang and Yin, Ming and Zheng, Boyuan and Yang, Zhenzhu and Liu, Yibo and Huang, Wenhao and Sun, Huan and Su, Yu and Chen, Wenhu},
  booktitle={Proceedings of the IEEE/CVF Conference on Computer Vision and Pattern Recognition (CVPR)},
  pages={9556--9567},
  year={2024},
  doi={10.1109/CVPR52733.2024.00913},
  url={https://openaccess.thecvf.com/content/CVPR2024/html/Yue_MMMU_A_Massive_Multi-discipline_Multimodal_Understanding_and_Reasoning_Benchmark_for_CVPR_2024_paper.html}
}

@inproceedings{jiang2020hover,
  title={{H}o{V}er: A Dataset for Many-Hop Fact Extraction And Claim Verification},
  author={Jiang, Yichen and Bordia, Shikha and Zhong, Zheng and Dognin, Charles and Singh, Maneesh and Bansal, Mohit},
  booktitle={Findings of the Association for Computational Linguistics: EMNLP 2020},
  pages={3441--3460},
  publisher={Association for Computational Linguistics},
  year={2020},
  doi={10.18653/v1/2020.findings-emnlp.309},
  url={https://aclanthology.org/2020.findings-emnlp.309/}
}

@inproceedings{chi2026evocf,
  title={Evo{CF}: Multi-Agent Collaboration via Agentic Memory-Driven Evolutionary Counterfactual Planning},
  author={Chi, Haotian and Feng, Zeyu and Yu, Xingrui and Luo, Linbo and Ong, Yew-Soon and Tsang, Ivor and Chen, Hechang and Chang, Yi and Yin, Haiyan},
  booktitle={Proceedings of the 43rd International Conference on Machine Learning},
  series={Proceedings of Machine Learning Research},
  volume={306},
  year={2026},
  url={https://openreview.net/forum?id=FsSirPJy9U}
}

@inproceedings{zhou2023contextfaithful,
  title={Context-faithful Prompting for Large Language Models},
  author={Zhou, Wenxuan and Zhang, Sheng and Poon, Hoifung and Chen, Muhao},
  booktitle={Findings of the Association for Computational Linguistics: EMNLP 2023},
  pages={14544--14556},
  publisher={Association for Computational Linguistics},
  year={2023},
  doi={10.18653/v1/2023.findings-emnlp.968},
  url={https://aclanthology.org/2023.findings-emnlp.968/}
}
\bibliographystyle{iclr2027_conference}

\newpage
\appendix

\vspace{1cm}
\centering
\textbf{\Large{Supplementary Material for 
``Counterfactual Self-Evolving Agents for Evidence Grounded Reasoning''}}

\justifying
\setlength{\parindent}{0pt}
\vspace{0.5cm}

\section{Extended Related Works} \label{app:related}

\subsection{Feedback and Memory for Self-Improvement}
A central distinction in self-improvement is what changes after feedback: an answer, stored
experience, an executable program, or model parameters. Output refinement and episodic memory
use language as the feedback medium. Self-Refine revises a current answer through self-critique
\citep{madaan2023selfrefine}, while Reflexion carries verbal reflections into later attempts
\citep{shinn2023reflexion}. In clinical reasoning, GSEM organizes experience into a graph and
calibrates retrieval through outcome-dependent updates to node quality and edge weights
\citep{han2026gsem}. These mechanisms make feedback reusable, but feedback quality remains
central: retaining a critique does not establish that it identifies an error in the original
decision. Our Proposer is therefore trained to connect an evidence edit to a hypothesized
mechanism and to assess whether the original evidence warrants correction or preservation.\looseness=-1

Externally evaluable tasks provide stronger grounds for accepting updates. Code evolution and
collaborative discovery use coding benchmarks or separate task evaluators to assess candidate
solutions, as in the Darwin G\"odel Machine and CORAL
\citep{zhang2025dgm,qu2026coral}. Propose, Solve, Verify uses formal verification to support
self-play between a problem proposer and solver \citep{wilf2025propose}. Our clinical Verifier
instead checks evidence consistency and plausibility using retrieved knowledge; its judgments
are not formal correctness certificates. We train the Proposer using Solver and Verifier
feedback, then evolve memory while keeping agent parameters fixed. This separates learning to
produce useful feedback from accumulating accepted experience.

\subsection{Counterfactual Objectives and Agent Evolution}
Counterfactual generation serves different objectives depending on what is altered and how the
result is used. For model explanation, FIZLE uses localized text edits to probe black-box
classifiers \citep{bhattacharjee2024fizle}. For robust learning, CATO uses causal assumptions
to guide augmentation that varies spurious attributes while preserving the target
\citep{feder2023augmentations}. SenseCF learns to generate health-related counterfactuals for
intervention design and sensor-data augmentation \citep{soumma2026counterfactual}. These uses
establish the value of controlled changes, but evaluating an altered input answers a different
question from deciding whether the original prediction should be revised. Our instruction
targets require the Proposer to return to the original case and examine the support that remains,
rather than treating a prediction change alone as evidence of improvement.

Counterfactuals can also guide search and select experience for learning. EvoCF induces
constraints from execution failures, generates alternative multi-agent plans through
counterfactual mutations, and evaluates candidates using retrieved experience
\citep{chi2026evocf}. It provides a close precedent for combining counterfactual search with
persistent memory, with plans as the object of intervention. I-EDI instead tests reasoning
traces through executable counterfactual perturbations and structural consistency checks,
then distills accepted traces into the solver \citep{fan2026iedi}. Our interventions concern
case evidence and its interpretation; the learned component is the Proposer, while the Solver
receives accepted counterfactual feedback as context. The distinction is thus the adaptation
target and the requirement to justify correction or preservation on the original case, rather
than counterfactual generation or memory alone.

\subsection{Evidence Use in Clinical Reasoning}
Clinical agent systems improve evidence use through complementary forms of coordination.
Role-based discussion and adaptive team formation organize specialist perspectives
\citep{tang2024medagents,kim2024mdagents}; guideline retrieval and multimodal tools connect
reasoning to external clinical evidence \citep{wang2026medagentpro}. Counterfactual diagnostic
discussion makes the dependence on individual findings more explicit:
\citet{you2026cfmultiagent} edit findings and use resulting confidence differences to guide
specialist deliberation and revise diagnoses. This training-free framework is particularly
relevant to our within-case feedback loop. Our additional focus is learning the Proposer's
feedback policy and retaining accepted experience for subsequent cases.

Evidence representation is a separate challenge in irregular clinical records. FuseMoE
integrates missing and irregularly sampled modalities through expert routing
\citep{han2024fusemoe}, whereas MILM serializes observations as time-ordered triplets and uses
staged training to learn sampling patterns before incorporating measured values
\citep{chung2026milm}. These methods motivate our predictive baselines and the need to preserve
temporal and missingness information. Complementary advances improve attention and expert
routing: quadratic-gating theory motivates nonlinear transformations of the attention value
matrix \citep{akbarian2024quadratic}, and hierarchical MoE analysis motivates Laplace gating
to accelerate expert estimation \citep{nguyen2024hierarchical}. For multimodal representations,
MERGE guides specialized experts using temporal redundancy, uniqueness, and synergy between
modalities \citep{han2026merge}. Our contribution concerns how a Solver examines and
reuses such evidence through counterfactual feedback, complementing improvements to the
underlying representation.

\section{Additional Experimental Results}

\subsection{Experimental Setup for MCQ Evaluations}
\label{app:mcq-setup}

This section describes DiagnosisArena and MMMU Business. Each dataset--model
pair uses a separately trained Proposer and a frozen Solver. Clinical
IHM/LOS training is described in Section~\ref{app:cf-training}.

\textbf{Data and validation.}
DiagnosisArena contains 915 four-choice diagnosis questions. MMMU Business
contains 967 strict A--D questions from Accounting, Economics, Finance,
Management, and Marketing. Table~\ref{tab:mcq-benchmark-splits} gives the
benchmark splits. Supervision constructed from training questions is
further divided into Proposer fitting and internal-validation records.
Internal validation is used for checkpoint selection; the external
benchmark validation set is used for CF-acceptance settings. Test questions
are excluded from both fitting sets. One supervision record can contain
multiple CFs, so record counts differ from CF and atomic-edit counts.

\begin{table}[!htbp]
\centering
\begin{minipage}{0.78\linewidth}
\centering
\caption{\textbf{Benchmark splits for MCQ evaluation.}}
\label{tab:mcq-benchmark-splits}
\small
\setlength{\tabcolsep}{3pt}
\renewcommand{\arraystretch}{1.10}
\begin{tabular*}{\linewidth}{@{\extracolsep{\fill}}lrrr@{}}
\toprule
Dataset & Training & Validation & Test \\
\midrule
DiagnosisArena & 640 & 137 & 138 \\
MMMU Business & 677 & 145 & 145 \\
\bottomrule
\end{tabular*}
\end{minipage}
\end{table}

\textbf{Image and Proposer inputs.}
For MMMU Business, the recorded Qwen3.6-27B captioning configuration converts
images into fixed descriptions of visible text, numbers, tables, axes,
and spatial relations without reference answers. The text-based Solvers
receive these descriptions rather than raw images. At inference, each
Proposer receives the question, options, and its Solver's initial answer
and reasoning. DiagnosisArena additionally supplies the top-two logit
margin and an evidence catalog with IDs and exact source text, to describe
decision uncertainty and locate executable edits. Reference answers,
record IDs, reference CFs, and post-CF Solver answers are excluded from
Proposer inputs. Humans design the CF supervision and human experts verify
it, with Claude Opus 4.6 assisting verification through AWS Bedrock. Training
answers support this offline review and are excluded at inference. Solver
and Verifier parameters remain frozen.

\textbf{Outcome measures.}
Accuracy is exact A--D match over the full test set. We also count
\emph{corrections} (wrong to correct) and \emph{harmful revisions}
(correct to wrong), because changing an answer is not necessarily useful.
For $B$ initially correct answers, $C$ corrections, $H$ harmful revisions,
and $n$ test questions,\looseness=-1
\[
N_{\mathrm{correct,final}}=B+C-H,\qquad
\Delta_{\mathrm{pp}}=100(C-H)/n.
\]
The correction rate is $C/(n-B)$ and preservation is $(B-H)/B$, when the
denominators are nonzero. Changes between two wrong options are counted
separately, and valid-output counts are reported where available. Paired
statistics use the initial predictions recorded in the same full-system
run as the revised predictions, including its output-processing protocol.
A separate zero-shot result need not be that run's starting point.

\textbf{Comparison scope.}
Cross-model evaluation tests the framework with a separate Proposer and
retrieval configuration for each base model, rather than zero-shot
transfer of one shared adapter. Seed averages and best recorded complete
runs are identified separately in the relevant results.
\FloatBarrier

\subsection{Full IHM Component Comparison (RQ1)}
\label{app:rq1-cfit-full}
Table~\ref{tab:app-rq1-cfit-full} reports all four IHM metrics for the component
comparison summarized in Figure~\ref{tab:rq1_cfit_main}(a). CF instruction
tuning is implemented as supervised fine-tuning (SFT). The SFT Proposer is
trained on 3,000 training CFs; SFT+GRPO is its step-64 continuation
after one RL epoch. These parameter updates precede the memory-evolution
rounds, during which both models remain frozen. ``Qwen SFT'' denotes a
task-fine-tuned Solver. The zero-shot, SFT, and SFT+GRPO runs share all 10,471
IHM test cases, the frozen Qwen Solver, and the evaluation implementation.
Appendix~\ref{app:collective-feedback} specifies the reported reward.

\begin{table}[!htbp]
\centering
\small
\setlength{\tabcolsep}{5.5pt}
\caption{\textbf{Component comparison on full-test IHM (RQ1).} SFT Proposer and
SFT+GRPO are evaluated with the frozen Qwen2.5-7B Solver; the Solver-specific row uses a
Proposer conditioned on the SFT Solver's draft. Higher is better.}
\label{tab:app-rq1-cfit-full}
\begin{tabular}{llrrrr}
\toprule
Proposer & Solver & Accuracy & AUROC & AUPRC & F1 \\
\midrule
None (zero-shot) & Qwen & .5839 & .7000 & .2252 & .3052 \\
SFT & Qwen & .7372 & .7543 & .2985 & \textbf{.3723} \\
None & Qwen SFT & \textbf{.8786} & .7584 & .3128 & .1216 \\
Solver-specific SFT & Qwen SFT & .8630 & \textbf{.7595} & .3212 & .3521 \\
SFT + GRPO & Qwen & .8417 & .7553 & \textbf{.3237} & .3720 \\
\bottomrule
\end{tabular}
\end{table}

\subsection{DiagnosisArena Instruction Tuning and Training Scale}
\label{app:diag-cf-it}
\label{app:rq1_cf_training}

\textbf{Learned guidance improves the frozen Solver.}
Experiment~4 evaluates a Proposer trained on human-designed, expert-verified
CF supervision with Claude-assisted verification. The frozen
Qwen2.5-7B-Instruct Solver initially answers
48 of 138 test questions correctly (34.8\%). The three Proposer runs reach
60.9\%, 63.0\%, and 58.7\%, with 34--41 corrections and 0--2 harmful
revisions each (Table~\ref{tab:app-diag-cf-it}). Mean accuracy is 60.9\%
with a sample standard deviation of 2.2 percentage points. All three
reported seeds improve the frozen Solver, supporting learned CF guidance
as a route to better decisions without updating Solver parameters.

\begin{table}[!htbp]
\centering
\begin{minipage}{0.92\linewidth}
\centering
\caption{\textbf{DiagnosisArena CF instruction-tuning results.}}
\label{tab:app-diag-cf-it}
\small
\setlength{\tabcolsep}{3pt}
\renewcommand{\arraystretch}{1.10}
\begin{tabular*}{\linewidth}{@{\extracolsep{\fill}}lrrrr@{}}
\toprule
Configuration & \shortstack{Test\\correct} & \shortstack{Accuracy\\(\%)} & Corrected & Harmed \\
\midrule
Zero-shot Solver & 48 & 34.8 & --- & --- \\
CF-IT, seed 20260728 & 84 & 60.9 & 36 & 0 \\
CF-IT, seed 20260730 & 87 & 63.0 & 41 & 2 \\
CF-IT, seed 20260731 & 81 & 58.7 & 34 & 1 \\
\midrule
Three-seed mean & 84.0 & $60.9\pm2.2$ & 37.0 & 1.0 \\
Validation-selected consensus & 87 & \textbf{63.0} & 39 & \textbf{0} \\
\bottomrule
\end{tabular*}
\end{minipage}
\end{table}

\textbf{Consensus retains corrections with fewer harmful changes.}
All values in Table~\ref{tab:app-diag-cf-it} are test results. CF-IT means
CF instruction tuning; the mean and standard deviation use only the three
individual runs. The consensus combines their outputs using a rule
selected on the external 137-question validation set and frozen before
testing. It reaches 87/138 correct answers by correcting 39 of the
90 initially wrong answers and preserving all 48 initially correct ones.
Seed 20260730 reaches the same accuracy with 41 corrections and two harms.
Thus, equal accuracy can reflect different correction--preservation
balances. Construction and training details appear in
Section~\ref{app:mcq-cf-construction}.

\textbf{Training scale and preservation.}
Experiment~5 uses nested subsets of 133, 266, 400, and 533 records,
stratified by \texttt{CROSSED}/\texttt{NOT\_CROSSED}, with the same
93-record internal-validation set and a fixed five-epoch recipe.
Table~\ref{tab:app-diag-cf-scale} reports one completed run per size,
not a seed average. All outcome columns refer to the same 138-question
test set, starting from 48 correct answers. The last column counts
switches from one wrong option to another.

\begin{table}[!htbp]
\centering
\begin{minipage}{0.86\linewidth}
\centering
\caption{\textbf{DiagnosisArena training-scale results.}}
\label{tab:app-diag-cf-scale}
\small
\setlength{\tabcolsep}{3pt}
\renewcommand{\arraystretch}{1.10}
\begin{tabular*}{\linewidth}{@{\extracolsep{\fill}}rrrrrr@{}}
\toprule
\shortstack{Training\\records} & \shortstack{Test\\correct} &
\shortstack{Accuracy\\(\%)} & Corrected & Harmed &
\shortstack{Different\\wrong option} \\
\midrule
133 & 48 & 34.8 & 1 & 1 & 0 \\
266 & 48 & 34.8 & 0 & 0 & 0 \\
400 & 77 & 55.8 & 36 & 7 & 6 \\
533 & 84 & \textbf{60.9} & 36 & \textbf{0} & 1 \\
\bottomrule
\end{tabular*}
\end{minipage}
\end{table}

The 133- and 266-record runs remain at 34.8\% accuracy. The 400- and
533-record runs reach 55.8\% and 60.9\%. Both correct 36 errors, but
harmful revisions fall from seven to zero. Their 5.1-percentage-point
difference therefore comes from preserving more initially correct
answers, rather than a larger correction count. Fixed epochs also give
larger datasets more optimizer updates, so this is a comparison of overall
training scale under that recipe. The observed result supports evaluating
CF guidance through both correction and preservation.
\FloatBarrier

\section{Counterfactual Instruction Tuning}
\label{app:cf-it}

\subsection{Failure modes of early counterfactual construction}
\label{app:cf-failures}
Two early approaches motivated instruction tuning: heuristic construction could produce ungrounded edits, and flip-reward learning did not distinguish correction from harm. Their failures show why the Proposer must learn whether an evidence change warrants reconsideration before optimizing its effect on the Solver.

\begin{figure}[htbp]
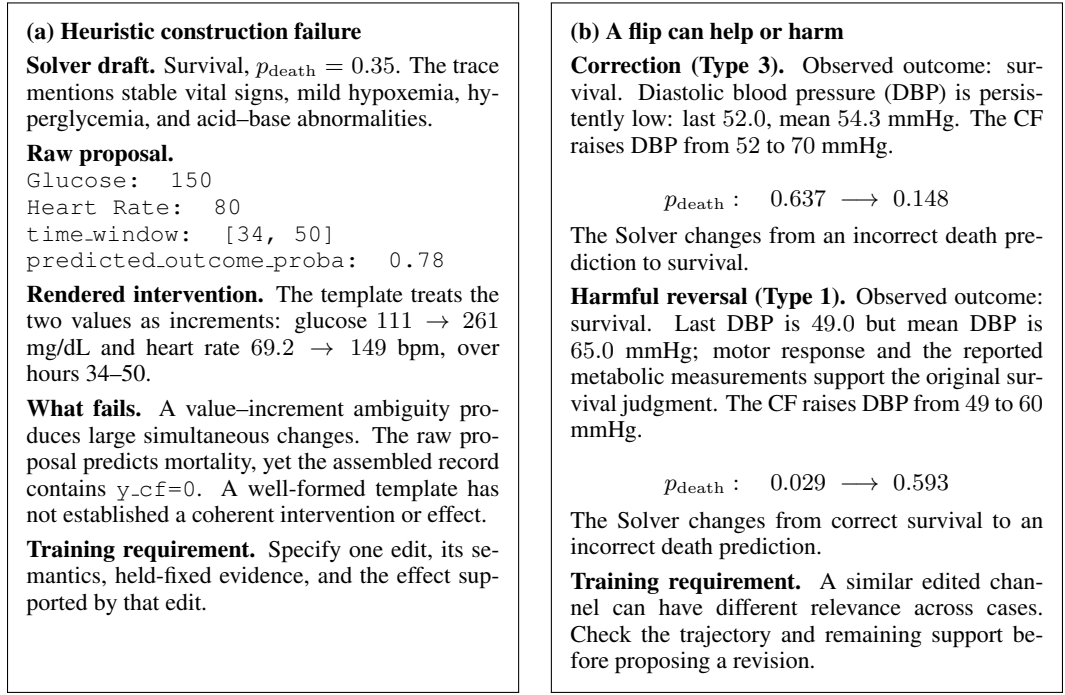

\centering
\begingroup
\setlength{\fboxsep}{7pt}
\begin{minipage}[t]{0.485\linewidth}
\vspace{0pt}\fbox{\begin{minipage}[t][245pt][t]{\dimexpr\linewidth-2\fboxsep-2\fboxrule\relax}
\small
\textbf{(a) Heuristic construction failure}\par\smallskip
\textbf{Solver draft.} Survival, $p_{\mathrm{death}}=0.35$. The trace mentions stable vital signs, mild hypoxemia, hyperglycemia, and acid--base abnormalities.\par\smallskip
\textbf{Raw proposal.}\par
\texttt{Glucose: 150}\par
\texttt{Heart Rate: 80}\par
\texttt{time\_window: [34, 50]}\par
\texttt{predicted\_outcome\_proba: 0.78}\par\smallskip
\textbf{Rendered intervention.} The template treats the two values as increments: glucose $111\to261$ mg/dL and heart rate $69.2\to149$ bpm, over hours 34--50.\par\smallskip
\textbf{What fails.} A value--increment ambiguity produces large simultaneous changes. The raw proposal predicts mortality, yet the assembled record contains \texttt{y\_cf=0}. A well-formed template has not established a coherent intervention or effect.\par\smallskip
\textbf{Training requirement.} Specify one edit, its semantics, held-fixed evidence, and the effect supported by that edit.
\end{minipage}}
\end{minipage}\hfill
\begin{minipage}[t]{0.485\linewidth}
\vspace{0pt}\fbox{\begin{minipage}[t][245pt][t]{\dimexpr\linewidth-2\fboxsep-2\fboxrule\relax}
\small
\textbf{(b) A flip can help or harm}\par\smallskip
\textbf{Correction (Type 3).} Observed outcome: survival. Diastolic blood pressure (DBP) is persistently low: last $52.0$, mean $54.3$ mmHg. The CF raises DBP from $52$ to $70$ mmHg.\par
\[p_{\mathrm{death}}:\quad 0.637\;\longrightarrow\;0.148\]
The Solver changes from an incorrect death prediction to survival.\par\smallskip
\textbf{Harmful reversal (Type 1).} Observed outcome: survival. Last DBP is $49.0$ but mean DBP is $65.0$ mmHg; motor response and the reported metabolic measurements support the original survival judgment. The CF raises DBP from $49$ to $60$ mmHg.\par
\[p_{\mathrm{death}}:\quad 0.029\;\longrightarrow\;0.593\]
The Solver changes from correct survival to an incorrect death prediction.\par\smallskip
\textbf{Training requirement.} A similar edited channel can have different relevance across cases. Check the trajectory and remaining support before proposing a revision.
\end{minipage}}
\end{minipage}
\endgroup
\caption{\textbf{Early CF failures motivate grounded, selective revision.} (a) A heuristic proposal becomes an inconsistent intervention. (b) Two pilot cases show opposite accuracy effects despite both changing the Solver's decision. Probabilities refer to the original patient before and after CF-context injection, not measured outcomes of the hypothetical intervention.}
\label{fig:cf-failure-examples}
\end{figure}

The training-free baseline selected channels by matching keywords in the Solver's reasoning. For example, mentions of acidosis could select bicarbonate or anion gap, while mentions of infection could select white blood cells. When the trace supplied no time interval, the heuristic used the final third of the available trajectory. The selected channels and interval were passed to the Proposer with a predefined output template. This procedure conflated how often a feature was mentioned with whether it was decisive, and left the meaning of numeric edits insufficiently constrained. Figure~\ref{fig:cf-failure-examples}a shows a resulting mismatch between the raw proposal and the rendered intervention. These examples belong to an early prototype that used longer trajectories; the current protocol restricts patient evidence to the prediction window.

The second failure concerns the learning objective. A reward for $\hat y^{+}\ne\hat y$ gives positive feedback to both a correction and a harmful reversal. In the early flip-reward RL attempt, these opposing transitions canceled the accuracy benefit. Figure~\ref{fig:cf-failure-examples}b gives two pilot examples of the underlying behavior. They illustrate the distinction in Equation~\ref{eq:correction-preservation}, rather than estimating transition frequencies or isolating the causal effect of RL training. The instruction targets therefore teach the Proposer to inspect residual support and to preserve a sound decision even when a plausible CF can be generated.

\subsection{Structural semantics of counterfactual edits}
\label{app:cf-scm}
Let $M=(U,V,F,P_U)$ be an acyclic structural causal model with equations $V_j=f_j(\mathrm{Pa}_j,U_j)$. A background realization $u$ determines one factual world. The intervened model $M_{a'}$ replaces only the equation for action $A$ with $A=a'$; other structural functions and the background realization remain unchanged. The notation $V_j^{\mathrm{do}(A=a')}(u)$ denotes the solution for $V_j$ in $M_{a'}$. This is the structural semantics of counterfactuals described by \citet[Chapter~7]{pearl2009causality}.

For observed evidence $E=e$ and a query variable $Z\in V$, abduction--action--prediction takes the form
\begin{equation}
  \Pr_M(Z_{a'}=z\mid E=e)
  =\int\mathbf{1}\{Z_{M_{a'}}(u)=z\}\,dP_M(u\mid E=e).
  \label{eq:cf-aap-semantics}
\end{equation}
Abduction updates the background distribution using the factual evidence, action replaces the designated mechanism, and prediction evaluates the query in the modified model. The same background realization couples the factual and hypothetical worlds. The Proposer's structured target follows this organization as a language-based audit; the present implementation does not estimate the functions $F$, the posterior $P_M(u\mid E=e)$, or numerical instance-level counterfactual outcomes from this equation.

\paragraph{Local preservation.}
Let $\mathrm{Desc}_M(A)$ denote the strict descendants of $A$. For every $V_j\notin\{A\}\cup\mathrm{Desc}_M(A)$, Equation~\ref{eq:cf-local-preservation} holds at every fixed $u$.

\noindent\textbf{Proof.}
Order the variables topologically. A variable outside $A$ and its descendants has no parent in that set, since such a parent would make it a descendant. Its structural equation is unchanged. Starting from the roots, its parents retain their factual values by induction, and its background inputs retain their values because $u$ is fixed. Applying the same function to the same inputs yields the same output. No independence assumption among the background variables is needed. Descendants may change when their inputs change, so replacing one mechanism does not imply changing only one observed value.

The three CF families act on different parts of this reasoning process. Observe changes the Solver's access to a fixed record; it can be represented by a visibility variable $B_j$ in $O_j=h_j(V_j,B_j)$, where $O_j$ is the exposed observation. Changing $B_j$ does not change the underlying state $V_j$. Explain compares hypotheses about latent causes using the same evidence, through quantities such as $P(H\mid E=e)$; selecting an alternative explanation is not automatically an intervention on $H$. Intervene changes an action mechanism in $M$ and permits downstream propagation. These distinctions specify the intended constraints for the \texttt{preserved}, \texttt{evidence\_delta}, and mechanism fields. They do not establish that a generated causal hypothesis is true; the decision analysis below evaluates its effect on the Solver's answer to the original case.

\subsection{Causal benefit and selective proposal}
\label{app:cf-causal-analysis}
We analyze the causal effect of supplying CF context to a frozen Solver. The unit is an original task instance with a fixed reference answer $y$. A fixed Proposer procedure generates a candidate from the available evidence and Solver draft, without access to $y$. The two procedures are $T=0$, retaining the original decision, and $T=1$, revising that decision after reading the candidate. Both answer the original question using the original record; the hypothetical clinical outcome is not the evaluation label. Thus the estimand concerns the effect of context on Solver correctness, not the effect of a clinical treatment on a patient.

Let $R_t=\mathbf{1}\{\hat y(t)=y\}$ be the potential correctness under procedure $t$. The Solver, Proposer, and memory state are fixed during this comparison, with no updates between evaluation cases. Deterministic execution defines both responses for each case. With stochastic execution, a specified joint randomization scheme defines their pairing; correction and harm probabilities refer to that scheme. Marginal accuracy differences do not require a unique pairing. Running both procedures on the same labeled cases measures the joint responses under the chosen protocol. Comparing different case sets would not establish that joint distribution. To isolate the CF's content from additional computation, an evaluation can further match the revision procedure with a neutral-context control.

\paragraph{Proposition 1: correction and harm from marginal accuracies.}
Write $p_t=\Pr(R_t=1)$, $b=\Pr(R_0=0,R_1=1)$, and $h=\Pr(R_0=1,R_1=0)$. Then Equation~\ref{eq:cf-benefit-bounds} gives the sharp bounds on $b$ and the corresponding value of $h$. Here $b$ is the probability that context provision is both necessary and sufficient for correctness, an application of the probabilities-of-causation framework of \citet{tian2000probabilities}. The bound is a specialization of their Equation~23, rather than a new general identification result. No monotonicity assumption that context cannot harm a correct answer is imposed.

\noindent\textbf{Proof.}
Let $s=\Pr(R_0=1,R_1=1)$. The joint response probabilities are
\begin{equation}
\begin{array}{c|cc}
 & R_1=0 & R_1=1 \\
\hline
R_0=0 & 1-p_0-p_1+s & p_1-s \\
R_0=1 & p_0-s & s
\end{array}.
\label{eq:cf-joint-responses}
\end{equation}
Nonnegativity requires
\begin{equation}
\max(0,p_0+p_1-1)\le s\le\min(p_0,p_1).
\end{equation}
Substituting $b=p_1-s$ yields the stated bounds, while $h=p_0-s=b-(p_1-p_0)$. Every $s$ in this interval defines a valid joint distribution, establishing sharpness. For example, $p_0=0.70$ and $p_1=0.80$ permit $b\in[0.10,0.30]$ and $h\in[0,0.20]$. These illustrative values show why a positive accuracy difference does not determine the harm rate. Paired evaluation additionally estimates the conditional correction rate $b/(1-p_0)$ and preservation rate $1-h/p_0$, when the denominators are nonzero.

The correctness-based decomposition applies to any finite answer space. Binary predictions satisfy $\Pr(\hat y(1)\ne\hat y(0))=b+h$. In MCQ tasks, there is an additional term:
\begin{equation}
\Pr(\hat y(1)\ne\hat y(0))
=b+h+\Pr(R_0=R_1=0,\,\hat y(1)\ne\hat y(0)).
\end{equation}
Switching between wrong options therefore increases the flip rate without improving accuracy, as illustrated in Figure~\ref{fig:cf-instruction-tuning}b. Strong/weak CF supervision addresses a different distinction: removing nonessential evidence should retain diagnostic support, whereas removing a diagnostic pillar should weaken it. The latter does not, by itself, identify a new correct option.

\paragraph{Proposition 2: optimal selective proposal.}
Let $X$ contain all information used by the gate before injection, including the available evidence, original Solver draft, and candidate CF, but excluding the reference answer. Define $b(X)=\Pr(R_0=0,R_1=1\mid X)$ and $h(X)=\Pr(R_0=1,R_1=0\mid X)$. Consider an inject-or-abstain policy $g(X)\in\{0,1\}$, where abstention retains the original decision exactly. Relative to abstention, let correction earn $\kappa_+>0$, harm cost $\kappa_->0$, and unchanged correctness have zero incremental utility. Among policies using $X$, Equation~\ref{eq:cf-selective-rule} maximizes expected incremental utility, with ties resolved by abstention.

\noindent\textbf{Proof.}
Conditioning on $X$ gives
\begin{equation}
\mathbb{E}[U_g\mid X]
=g(X)\left[\kappa_+b(X)-\kappa_-h(X)\right].
\end{equation}
The maximizing choice is $g(X)=1$ exactly when the bracket is positive; taking expectations preserves this pointwise optimum. For equal weights, define $\tau(X)=b(X)-h(X)$. The gated correctness satisfies $R_g=(1-g(X))R_0+g(X)R_1$, hence
\begin{equation}
\mathbb{E}[R_{g^*}]-\mathbb{E}[R_0]
=\mathbb{E}[\max\{\tau(X),0\}]\ge0.
\label{eq:cf-ideal-gate-gain}
\end{equation}
This is a population statement under the true conditional response probabilities, not a guarantee for an estimated gate or for each individual case.

To expose the quantities a gate must assess, write
\begin{equation}
\begin{aligned}
w(X)&=\Pr(R_0=0\mid X),\\
\alpha(X)&=\Pr(R_1=1\mid R_0=0,X),\\
\beta(X)&=\Pr(R_1=0\mid R_0=1,X).
\end{aligned}
\end{equation}
Then $b(X)=w(X)\alpha(X)$ and $h(X)=[1-w(X)]\beta(X)$, so proposing is preferred when
\begin{equation}
\kappa_+w(X)\alpha(X)>\kappa_-[1-w(X)]\beta(X).
\end{equation}
Conditional probabilities within a zero-probability stratum can be assigned arbitrarily because their multiplier vanishes. Original-answer uncertainty alone does not determine this inequality: the prospect of correcting an error and the risk of damaging a correct decision also matter. The instruction targets teach these distinctions through corrective, preservation, and \texttt{no\_cf} examples. The implemented Proposer does not explicitly estimate all three probabilities, and its reported confidence is not assumed calibrated. Applying the guarantee in Equation~\ref{eq:cf-ideal-gate-gain} to a learned gate would require additional estimation and validation. Training labels may supervise these response types, but they remain unavailable to the Proposer at inference.

\subsection{Data construction and input boundaries}
\label{app:cf-data}
The current clinical input serializes the first 24 hours of each patient record into baseline demographics and history, physiological summaries, support and medication states, outputs, devices, transitions, and compressed radiology findings available before the cutoff. Unobserved channels are omitted, empty sections indicate that no information is available, and singleton measurements retain their observation count. The Proposer then receives the target Solver's discrete prediction, continuous probability, and full reasoning trace, followed by the CF audit request. The fields \texttt{y\_true}, \texttt{planner\_correct}, \texttt{teacher\_signal}, and \texttt{hospital\_expire\_flag} are excluded at prompt construction.

Training cases cover errors in evidence use, including omitted findings, incorrect numerical readings, temporal order, support state, and relationships between findings. Corrective, hard preservation, and anchor preservation records are selected from different training stays with related phenotypes or error mechanisms. Matching selects relevant cases; it does not permit copying another patient's measurements into a target. All factual numbers must be traceable to the current patient input, and hypothetical changes must be marked explicitly. If an observed outcome cannot be explained from the prediction-window evidence, the target expresses uncertainty or abstains instead of inventing a corrective mechanism. Data checks cover numerical grounding, hidden-outcome absence, unique stays, and separation of training and test patients. Development audits used test reasoning to identify abstract error types; test patient facts and outcomes were not copied into instruction-tuning messages. Test labels were reserved for scoring, but test reasoning was therefore not wholly unexamined during development.

The XML contract supports six operation values. \texttt{action\_change} modifies a treatment or support state; \texttt{temporal\_shift} changes an event's timing or duration; \texttt{cause\_swap} substitutes a hypothesized explanation of an observed phenomenon; \texttt{modality\_ablation} changes access to an information channel; \texttt{evidence\_addition} introduces an explicitly hypothetical finding; and \texttt{no\_cf} abstains. These operations instantiate the three conceptual families in Section~\ref{sec:cf-instruction-tuning}; the family of a timing edit depends on whether it changes the clinical process or only access to evidence. A non-abstaining target names the operation, states a single local change, identifies the unaffected evidence paths, explains the mechanism, and returns to the actual case to assess whether a revision is warranted. The Solver-visible memory contains only the relevant current facts, the relation to reconsider, and its applicability boundary. It remains empty for abstention.

\subsection{Structured causal reasoning target}
\label{app:cf-target}
\begin{figure}[htbp]
\centering
\includegraphics[width=0.62\linewidth]{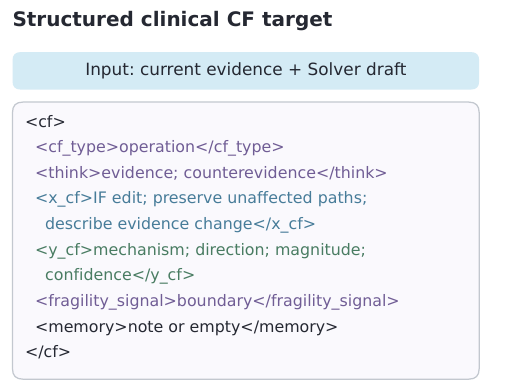}
\caption{\textbf{Structured clinical CF instruction target.} Current evidence and the original Solver draft condition an evidence audit, a hypothetical edit with unaffected paths held fixed, its expected effect, and an applicability boundary. Optional memory returns to the original case; abstention leaves it empty. The schema is condensed and is not a complete training record.}
\label{fig:cf-target-schema}
\end{figure}

The target follows the abduction--action--prediction organization of structural counterfactual reasoning \citep[Chapter~7]{pearl2009causality}, adapted to a language-based evidence audit. In \texttt{<think>}, the Proposer first reconstructs the explanation linking the observed evidence to the Solver's decision, locates an error or fragile premise, and considers counterevidence. This is the abductive step; a plausible explanation is kept distinct from an observed fact. In \texttt{<x\_cf>}, it specifies one patient-specific hypothetical change, the evidence held fixed, and the resulting local difference. For an action intervention, only unaffected paths are held fixed; downstream changes belong to the hypothesized effect. In \texttt{<y\_cf>}, it traces that effect and states the direction, magnitude, and confidence of the change in decision support, without assigning a flipped label. Finally, \texttt{<fragility\_signal>} compares the affected premise with residual support and states where the argument stops applying. The Proposer returns to the actual record before writing an optional \texttt{<memory>} note for the Solver. Abstention leaves memory empty. Figure~\ref{fig:cf-target-schema} summarizes the format; Appendix~\ref{app:cf-examples} gives corrective and preservation examples.

\subsection{Corrective and preservation target examples}
\label{app:cf-examples}
Figure~\ref{fig:cf-target-examples} gives condensed excerpts from the prototype instruction targets. The corrective example audits a factual error in the original draft; the preservation example tests whether removing one evidence channel leaves enough support. These excerpts illustrate the intended training behavior and do not report post-CF Solver performance. The corrective prototype used a 48-hour record, whereas the current clinical configuration uses the 24-hour input boundary described above.

\begin{figure}[htbp]
\centering
\begingroup
\setlength{\fboxsep}{7pt}
\begin{minipage}[t]{0.485\linewidth}
\vspace{0pt}\fbox{\begin{minipage}[t][223pt][t]{\dimexpr\linewidth-2\fboxsep-2\fboxrule\relax}
\small
\textbf{(a) Corrective target (Type 3)}\par\smallskip
\texttt{<think>}\par
The draft attributes high mortality risk to mean MAP $56.2$ mmHg. The record instead gives mean MAP $79.1$ and last MAP $92$ mmHg. Creatinine is $3.7\to3.7\to3.6$ mg/dL, and GCS motor response remains $6$. The low-mean-MAP premise is a factual misreading.\par\smallskip
\texttt{<y\_cf>}\par
Correcting that premise weakens the draft's mortality argument. Reassess the contribution of the remaining findings before changing the decision.\par\smallskip
\texttt{<fragility\_signal>}\par
The original argument relies heavily on the misread value. A stable creatinine sequence must remain an observation, not be rewritten as an unobserved diagnosis.\par\smallskip
\texttt{<memory>}\par
Recheck the cited MAP against the actual mean and latest value; reconsider the shock argument using the current record.
\end{minipage}}
\end{minipage}\hfill
\begin{minipage}[t]{0.485\linewidth}
\vspace{0pt}\fbox{\begin{minipage}[t][223pt][t]{\dimexpr\linewidth-2\fboxsep-2\fboxrule\relax}
\small
\textbf{(b) Preservation target (Type 2)}\par\smallskip
\texttt{<think>}\par
The survival judgment is supported by neurological recovery, extubation at hour 17, formal spinal clearance, stable hemodynamics, and an unremarkable metabolic panel.\par\smallskip
\texttt{<x\_cf>}\par
IF the GCS trajectory were unavailable, preserve the extubation, clearance, hemodynamic, and laboratory observations. Only neurological visibility changes.\par\smallskip
\texttt{<y\_cf>}\par
Removing this channel leaves several observations supporting survival. In this target, the ablation has little effect on the original decision's support.\par\smallskip
\texttt{<fragility\_signal>}\par
Robust to the proposed ablation; the remaining support does not warrant a reversal.\par\smallskip
\texttt{<memory></memory>}\par
Preserve the decision; inject no correction.
\end{minipage}}
\end{minipage}
\endgroup
\caption{\textbf{Complementary instruction targets.} (a) A corrective reasoning excerpt identifies a misread measurement. (b) A preservation excerpt tests evidence removal and retains the original decision. The examples are condensed and omit fields not needed to show the contrast; they are not complete training records.}
\label{fig:cf-target-examples}
\end{figure}

An action-based example in the current training corpus uses a different reasoning test. A patient with worsening hypoxemia receives escalating ventilatory support: PEEP increases from $5$ to $12$ cmH$_2$O and inspired oxygen from $0.50$ to $0.60$, while PaO$_2$ changes only from $43$ to $44$ mmHg. The original Solver predicts survival with death probability $0.35$. The CF asks what would happen without the escalation, holding the pre-intervention state fixed, then returns to the observed lack of improvement to challenge the Solver's treatment of support as reassurance. A complementary preservation case has brief postoperative vasopressor use, lactate decreasing from $3.3$ to $1.0$ mmol/L, and improving oxygenation. Its target preserves the original survival prediction rather than treating any vasopressor exposure as evidence of ongoing deterioration. Both examples come from training patients, and no hypothetical outcome is presented as observed.

\subsection{Optimization and training checks}
\label{app:cf-training}
IHM and LOS use separate task-specific adapters and training records with the same Qwen2.5-7B-Instruct backbone. Table~\ref{tab:cf-it-config} reports the documented full-scale clinical configuration. The Qwen chat template places the task contract in the system turn, patient evidence and the original Solver draft in the user turn, and the complete XML target followed by EOS in the assistant turn. Prompt and padding labels are masked with $-100$. Every remaining assistant token contributes equally to Equation~\ref{eq:cf-instruction-loss}; there is no separate loss on \texttt{x\_cf} or extra weighting of reasoning spans in this configuration.

For input $u_i$ and target token sequence $z_i$, we minimize the assistant-only autoregressive loss
\begin{equation}
  \mathcal{L}_{\mathrm{CF\text{-}IT}}(\theta)
  = -\frac{1}{\sum_i |z_i|}
    \sum_i\sum_{t=1}^{|z_i|}
    \log p_{\theta}(z_{i,t}\mid u_i,z_{i,<t}).
  \label{eq:cf-instruction-loss}
\end{equation}

\begin{table}[htbp]
\centering
\small
\begin{tabular}{p{0.33\linewidth}p{0.59\linewidth}}
\hline
Setting & Value \\
\hline
Base model & Qwen2.5-7B-Instruct \\
Training scale & $3{,}000$ records; $2$ epochs \\
Adapter & LoRA rank $16$, scaling $32$, dropout $0.05$ \\
Target projections & Attention: \texttt{q, k, v, o}; MLP: \texttt{gate, up, down} \\
Trainable parameters & $40{,}370{,}176$ \\
Learning rate and schedule & $10^{-4}$; $3\%$ warm-up; cosine decay \\
Effective batch size & $16$ (device batch $1$, accumulation $16$) \\
Sequence budget & $4{,}096$ tokens, retaining the complete XML target \\
Precision and memory & BF16; gradient checkpointing \\
Checkpointing & Save each epoch; retain the final adapter \\
Frozen components & Base backbone and Solver \\
\hline
\end{tabular}
\caption{Full-scale clinical Proposer instruction-tuning settings.}
\label{tab:cf-it-config}
\end{table}

Before training, checks confirm the intended projection coverage, trainable parameter count, and complete target visibility. Training-case replay then checks XML completeness, termination, factual grounding, correction versus preservation, and empty memory for \texttt{no\_cf}. Replay verifies that the interface has been learned; generalization is assessed separately on held-out cases. Earlier experiments also considered span-weighted assistant losses, but these are distinct from the uniform objective reported here.

\subsection{MCQ Supervision and Executable Counterfactuals}
\label{app:mcq-cf-construction}

The MCQ implementation trains the Proposer to inspect its Solver's
reasoning, propose a local hypothetical edit, and explain the resulting
decision boundary. It uses executable JSON targets, separately from the
clinical XML configuration above.

\subsubsection{Training labels and CF roles}

Offline, the frozen Solver answers each training question. Correct initial
answers are assigned \texttt{NOT\_CROSSED} (Type~2) for preservation
supervision; incorrect answers are assigned \texttt{CROSSED} (Type~3)
for corrective supervision. These labels describe training roles.
Actual correction and harm are measured from downstream paired answers.
At inference, the Proposer predicts its verdict without reference answers.

Table~\ref{tab:mcq-cf-roles} distinguishes intended CF effects. A
\emph{sufficient corrective} CF addresses decisive evidence to make the
reference answer clearly best; an \emph{insufficient corrective} CF concerns
the same boundary but leaves the answer non-unique or the case undetermined.
Both start from an incorrect Solver answer, whereas a safe-preserving CF
retains support for an already correct answer. Human experts verify these
sufficiency judgments with Claude assistance. They describe corrective targets,
separately from Figure~\ref{fig:cf-instruction-tuning}'s Strong/Weak ablations,
which test whether diagnostic support survives evidence removal.

\begin{table}[!htbp]
\centering
\begin{minipage}{0.82\linewidth}
\centering
\caption{\textbf{CF roles in MCQ supervision.}}
\label{tab:mcq-cf-roles}
\small
\setlength{\tabcolsep}{3pt}
\renewcommand{\arraystretch}{1.10}
\begin{tabularx}{\linewidth}{@{}>{\raggedright\arraybackslash}p{0.23\linewidth}>{\raggedright\arraybackslash}p{0.16\linewidth}>{\raggedright\arraybackslash}X@{}}
\toprule
CF role & Initial answer & Intended effect \\
\midrule
Safe-preserving & Correct & Preserve support for the reference answer after a non-decisive evidence change. \\
\addlinespace[2pt]
Sufficient corrective & Incorrect & Address the key reasoning error and make the reference answer clearly best. \\
\addlinespace[2pt]
Insufficient corrective & Incorrect & Address the same diagnostic boundary while leaving the reference answer non-unique. \\
\bottomrule
\end{tabularx}
\end{minipage}
\end{table}

\subsubsection{Human design and expert verification}

\textbf{Candidate design.}
Humans design the CF cases, and human experts verify their correctness.
Claude Opus 4.6, accessed through AWS Bedrock, assists verification. Each
case includes the original evidence, A--D options, Solver answer and
reasoning, and the training reference answer. The canonical DiagnosisArena
candidate pool contains five safe-preserving CFs for Type~2, or three
sufficient-intent and two insufficient-intent corrective CFs for Type~3.
Each candidate explains the relevant evidence and diagnostic boundary;
the contrast concerns diagnostic sufficiency, not wording or length.

\textbf{Apply, review, and select.}
Each candidate is mechanically applied to a fresh copy of the original
case. Expert review, assisted by Claude, checks medical coherence, role
consistency, answer leakage, grammar, contradictions, and grounding in
the edited case. The verification record includes decisive evidence spans
and execution checks. Accepted supervision contains one safe-preserving
CF for Type~2 or one sufficient/insufficient corrective pair for Type~3;
records with no acceptable candidate are excluded. Final records are
assembled locally. Solver replay is not used to assign supervision roles.

\subsubsection{Executable edits and quality checks}

A CF targets one evidence concept using the operations in
Table~\ref{tab:mcq-edit-contract}. Multiple atomic edits are allowed when
the same fact is repeated or a directly conflicting statement must also
be updated. This preserves the intended local change while keeping the
edited record internally consistent.

\begin{table}[!htbp]
\centering
\begin{minipage}{0.80\linewidth}
\centering
\caption{\textbf{Executable CF edits in DiagnosisArena.}}
\label{tab:mcq-edit-contract}
\small
\setlength{\tabcolsep}{3pt}
\renewcommand{\arraystretch}{1.10}
\begin{tabularx}{\linewidth}{@{}>{\raggedright\arraybackslash}p{0.17\linewidth}>{\raggedright\arraybackslash}p{0.35\linewidth}>{\raggedright\arraybackslash}X@{}}
\toprule
Operation & Required fields & Effect \\
\midrule
\texttt{delete} & Exact \texttt{target\_span} & Remove the specified source text. \\
\addlinespace[2pt]
\texttt{add} & Exact \texttt{anchor\_span}, \texttt{position}, \texttt{new\_text} & Insert explicitly hypothetical evidence at the specified location. \\
\addlinespace[2pt]
\texttt{replace} & Exact \texttt{target\_span}, \texttt{new\_text} & Replace a finding with a hypothetical alternative. \\
\bottomrule
\end{tabularx}
\end{minipage}
\end{table}

Source spans must match the designated \texttt{case}, \texttt{exam}, or
\texttt{tests} section verbatim and uniquely. Options remain unchanged;
inserted text cannot reveal the reference diagnosis, option letter, or
answer-selection instruction. Mechanical checks cover schema, source
matches, execution, text integrity, and explicit leakage. Failed outputs
may be retried with recorded errors, while raw outputs, applied cases,
and review decisions are retained.

\textbf{Canonical corpus.}
Of 636 eligible records, 626 remain after construction and verification
(98.4\%). They comprise 533 fitting records (363 corrective,
170 preservation) and 93 internal-validation records (63 corrective,
30 preservation). The 426 corrective pairs and 200 safe-preserving CFs
yield 1,052 CF objects.

\subsubsection{Solver-specific CF supervision}

Each dataset--model pair uses its own Solver drafts: the same question
may need correction for one Solver and preservation for another.
Tables~\ref{tab:diagnosis-supervision-pools} and
\ref{tab:mmmu-supervision-pools} report the supervision pools before
oversampling. ``Paired Type~3'' is a subset of Type~3, not additional
records. ``Train / val.'' refers to Proposer fitting and internal
validation, not the external benchmark validation set.

\begin{table}[!htbp]
\centering
\begin{minipage}{0.90\linewidth}
\centering
\caption{\textbf{DiagnosisArena supervision pools.}}
\label{tab:diagnosis-supervision-pools}
\small
\setlength{\tabcolsep}{3pt}
\renewcommand{\arraystretch}{1.10}
\begin{tabular*}{\linewidth}{@{\extracolsep{\fill}}lrrrrr@{}}
\toprule
Base model & \shortstack{Type 2\\preserve} & \shortstack{Type 3\\correct} &
\shortstack{Paired\\Type 3} & Pool & \shortstack{Train /\\val.} \\
\midrule
Qwen2.5-7B & 200 & 426 & 426 & 626 & 533 / 93 \\
Llama3-8B & 204 & 428 & 324 & 632 & 569 / 63 \\
MedGemma-4B & 255 & 375 & 296 & 630 & 567 / 63 \\
\bottomrule
\end{tabular*}
\end{minipage}
\end{table}

DiagnosisArena Qwen uses complete sufficient/insufficient pairs for all
corrective records. Llama and MedGemma combine paired records with
supplementary human-designed, expert-verified examples, with Claude Sonnet
4.6 assisting verification through Bedrock. These examples supply one
primary corrective or safe-preserving CF and undergo local executable-edit
and leakage checks; they do not require a five-candidate pool. Paired
coverage is 324/428
Type~3 records for Llama and 296/375 for MedGemma.

\begin{table}[!htbp]
\centering
\begin{minipage}{0.82\linewidth}
\centering
\caption{\textbf{MMMU Business supervision pools.}}
\label{tab:mmmu-supervision-pools}
\small
\setlength{\tabcolsep}{3pt}
\renewcommand{\arraystretch}{1.10}
\begin{tabular*}{\linewidth}{@{\extracolsep{\fill}}lrrrr@{}}
\toprule
Base model & \shortstack{Type 2\\preserve} & \shortstack{Type 3\\correct} & Pool & \shortstack{Train /\\val.} \\
\midrule
Qwen2.5-7B & 352 & 181 & 533 & 480 / 53 \\
Llama3-8B & 216 & 228 & 444 & 400 / 44 \\
MedGemma-4B & 213 & 248 & 461 & 415 / 46 \\
\bottomrule
\end{tabular*}
\end{minipage}
\end{table}

Business uses one primary CF per record without a paired insufficient
contrast. Its supervision covers formulas, variable roles, units, time bases,
table or graph interpretation, and option mapping. Both corrective and
preservation targets remain part of every model's supervision.

\subsubsection{Proposer Output and Optimization}

The Proposer learns a structured target rather than just an answer letter
(Table~\ref{tab:mcq-proposer-fields}). A missing contrast is
\texttt{null}; \texttt{UNRESOLVED} provides an abstention option. Reference
answers, archive IDs, reference CFs, and post-CF Solver outputs are not
Proposer inputs. The generated \texttt{proposed\_answer} remains an
answer-supervised output target, not a supplied test answer.

\begin{table}[!htbp]
\centering
\begin{minipage}{0.88\linewidth}
\centering
\caption{\textbf{MCQ Proposer target fields.}}
\label{tab:mcq-proposer-fields}
\small
\setlength{\tabcolsep}{3pt}
\renewcommand{\arraystretch}{1.10}
\begin{tabularx}{\linewidth}{@{}>{\raggedright\arraybackslash}p{0.32\linewidth}>{\raggedright\arraybackslash}X@{}}
\toprule
Target field & Learning target \\
\midrule
\texttt{boundary\_verdict} & Predict \texttt{CROSSED}, \texttt{NOT\_CROSSED}, or \texttt{UNRESOLVED}. \\
\texttt{proposed\_answer} & Predict the proposed A--D answer. \\
\texttt{closest\_competitor} & Identify the competing option relevant to the evidence boundary. \\
\texttt{think} & Explain the reasoning error or retained support and cite relevant evidence IDs. \\
\texttt{x\_cf.primary},\newline \texttt{x\_cf.contrast} & Generate the primary executable CF and an optional insufficient contrast. \\
\texttt{y\_cf} & Explain the expected effects on decision support. \\
\texttt{fragility} & Assess the original decision's robustness to the proposed evidence change. \\
\bottomrule
\end{tabularx}
\end{minipage}
\end{table}

\textbf{DiagnosisArena optimization.}
The documented Qwen2.5-7B-Instruct recipe uses 4-bit NF4 QLoRA with double
quantization, rank 16, scaling 32, and dropout 0.08. Adapters cover
attention \texttt{q/k/v/o} and MLP \texttt{gate/up/down} projections.
The learning rate is $5\times10^{-5}$ with cosine decay. Micro-batch size
one and four gradient-accumulation steps give effective batch size four.
Training allows up to five epochs with early-stopping patience two;
the retained messages fit the 8,192-token budget without target truncation.

Prompt and padding labels are masked with $-100$, focusing supervision on
the assistant CF output. The MCQ loss weights are 10 for
\texttt{boundary\_verdict} and \texttt{proposed\_answer}, 6 for
\texttt{closest\_competitor}, 3 for \texttt{x\_cf}, 2 for
\texttt{y\_cf} and \texttt{fragility}, and 1.5 for \texttt{think}.
This emphasizes decision and edit fields while retaining reasoning
supervision. These MCQ settings are separate from the uniform clinical
XML loss above.

Table~\ref{tab:mcq-fit-manifests} reports the model-specific fitting
configurations. Training records count distinct records; the next column
includes repeats from oversampling, not newly constructed cases. The last
column is the stopping epoch, not necessarily the best checkpoint's epoch.
These early-stopped configurations are separate from Experiment~5's
fixed-five-epoch series.

\begin{table}[!htbp]
\centering
\begin{minipage}{0.82\linewidth}
\centering
\caption{\textbf{Model-specific Proposer training settings.}}
\label{tab:mcq-fit-manifests}
\small
\setlength{\tabcolsep}{3pt}
\renewcommand{\arraystretch}{1.10}
\begin{tabular*}{\linewidth}{@{\extracolsep{\fill}}lrrr@{}}
\toprule
Base model & \shortstack{Training\\records} & \shortstack{After\\oversampling} & \shortstack{Stopping\\epoch} \\
\midrule
\multicolumn{4}{@{}l}{\textit{DiagnosisArena}}\\
Qwen2.5-7B & 533 & 703 & 4 \\
Llama3-8B & 569 & 749 & 3 \\
MedGemma-4B & 567 & 567 & 3 \\
\addlinespace[4pt]
\multicolumn{4}{@{}l}{\textit{MMMU Business}}\\
Qwen2.5-7B & 480 & 648 & 3 \\
Llama3-8B & 400 & 400 & 3 \\
MedGemma-4B & 415 & 415 & 3 \\
\bottomrule
\end{tabular*}
\end{minipage}
\end{table}

\subsubsection{What Reaches the Solver}

\textbf{Strict edit replay.}
The CF is applied to the case, examination, or test text. The Solver
receives the edited sections and original A--D options, without the
Proposer's reasoning audit, proposed answer, expected answer, reference
answer, or record ID. This evaluates the Solver on the edited input.

\textbf{Full closed-loop evaluation.}
An accepted current-instance CF is converted into evidence-boundary
context for the original question. The frozen Solver uses this guidance
to reconsider its initial answer. Mechanical checks establish
executability, retrieval-based verification assesses domain support,
and the validation-selected policy controls adoption. Full-system
correction and harm counts measure this revision of the original answer,
not the offline sufficiency label.

\subsubsection{Examples of Counterfactual Supervision}

\textbf{Preservation.}
In one canonical DiagnosisArena example, a hypothetical replacement makes
an imaging finding more alarming: remodeling and thinning become
expansile remodeling with cortical destruction. The example retains the
original histological evidence, including rare mitoses and no cytologic
atypia, as support for preservation. The intended lesson is to weigh a
changed finding against the evidence that still supports the answer.

\textbf{Corrective sufficiency contrast.}
Another example concerns a Solver that underweights a \textit{KRIT1}
variant and familial brain hemorrhage. The sufficient CF makes inheritance
explicit and introduces hypothetical shared-variant evidence in relatives.
The insufficient CF changes only the lack of improvement after prolonged
compression-stocking use. Expert verification, assisted by Claude,
identifies the former as sufficient and the latter as relevant but
insufficient for the diagnostic boundary.
These are supervision examples; hypothetical findings are not added to
the factual record. Together, they illustrate the intended distinction
between justified reconsideration and preservation.
\FloatBarrier

\section{Feedback from Verifier and Proposer}
\label{app:collective-feedback}

\paragraph{Verifier construction and coverage.}
The clinical Verifier first checks target ranges and deterministic safety constraints, then extracts the proposed condition, intervention, and relevant evidence. A topic router selects prebuilt Chroma collections. The configured PubMed Central (PMC) collections cover ICU prognosis, hemodynamics and vasopressors, consciousness, infection, respiratory support, hematology, metabolism, cardiac rhythm, and renal function. Acid--base queries share the metabolic collection. Three additional collections contain DailyMed drug labels, Surviving Sepsis Campaign guidelines, and American College of Radiology appropriateness criteria. These sources cover clinical literature, drug safety, sepsis management, and imaging decisions; they are not an exhaustive knowledge base for every specialty or task.

The implementation embeds condition, intervention, outcome, and critical-feature queries with Qwen3-Embedding-0.6B. It retrieves up to four chunks per query and collection, deduplicates them, and limits the judge's evidence bundle to 24 items, retaining source identifiers. The Qwen2.5-7B judge cites retrieved evidence and returns \texttt{PASS} when the assessed claims are supported, \texttt{FLAG} for incomplete support without contradiction, and \texttt{REJECT} for contradicted claims. Intervention plausibility and a hypothesized outcome can be evaluated separately; a plausibility verdict does not establish the unobserved outcome as fact. The three-level plausibility score is $r_V=1$ for \texttt{PASS}, $0.5$ for \texttt{FLAG}, and $0$ for \texttt{REJECT}. Deterministic safety rejection blocks the edit, and the grounded verifier pipeline additionally penalizes unsafe or insufficiently supported proposals. These verifier-side checks are distinct from the downstream flip-warrant gate.

\paragraph{Memory organization and use.}
The bank stores typed factual-case, verified-CF, and gap records with task/evidence tags, source identity, and verifier metadata. Tag overlap and text similarity retrieve a small relevant subset. Before reuse, a note is reduced to a conditional mechanism, its applicability requirements, counterexamples, and provenance. The Solver prompt separates current observations, the original Solver argument, the current-instance CF review, and retrieved training experience. Donor measurements are not presented as current observations. The current-instance note can refer to observed facts but keeps hypothetical changes explicit. If no usable note passes the gate, the Solver receives a fixed instruction to retain its initial assessment. Memory retrieval and the current-instance reviewer note are distinct components; the latter can be used without a cross-instance bank.

\paragraph{Training reward and the reported configuration.}
The reported clinical RL run uses an additive outcome-guided reward with weights $(\lambda_C,\lambda_U,\lambda_V)=(2,1,0)$, without the correctness veto in Equation~\ref{eq:collective-feedback-objective}:
\begin{equation}
\begin{aligned}
 R_{\mathrm{reported}}
 &=2r_C+\mathrm{clip}\!\left(\log p^c(y)-\log p^0(y),-2,2\right)\\
 &\quad+0.2r_{\mathrm{direction}}+0.1r_{\mathrm{schema}}-r_{\mathrm{violation}}.
\end{aligned}
\label{eq:reported-rl-reward}
\end{equation}
For the before/after correctness pairs $(0,1)$, $(1,0)$, $(1,1)$, and $(0,0)$, $r_C$ equals $3$, $-4$, $0.75$, and $-1$, respectively. Probabilities are clipped to $[10^{-4},1-10^{-4}]$ before taking logarithms. The confidence term is a reduction in true-answer log loss, not an entropy bonus: becoming more confident in a wrong answer does not earn credit. In the binary clinical configuration, the decision threshold is $0.5$.

The direction indicator is one when the reviewer note contains an explicit \texttt{INCREASE}, \texttt{DECREASE}, or \texttt{MAINTAINED} instruction. It rewards an explicit format, not medical validity. The schema indicator requires nonempty reasoning, edit, effect, and memory fields with no prohibited outcome disclosure. A malformed or label-leaking output is replaced with the fixed retain note and incurs $r_{\mathrm{violation}}=2$. If the Solver probability cannot be parsed, the implementation falls back to $p^c=p^0$ and applies the same penalty. Otherwise the violation penalty is zero. Reference labels are kept in reward-side metadata and never supplied to the Proposer prompt.

The Proposer starts from its instruction-tuned checkpoint and samples four candidates per training instance. Each candidate is scored through the frozen Solver, and group-relative policy updates compare their rewards. The documented clinical run uses 512 training patients and 128 disjoint validation patients; test cases are excluded from RL training. The separate DiagnosisArena RL implementation retains the same correctness-transition rewards but uses a clipped change in the correct-option logit margin for confidence and task-specific JSON/edit validity checks.

\subsection{MCQ RAG Verifier and Model-Specific Indexes}
\label{app:mcq-rag-verifier}

\textbf{Role and inputs.}
Retrieval serves different roles in Experiment~8: the retrieval-only
baseline supplies context directly to the Solver, whereas the retrieval
stage described here supplies evidence to a Verifier assessing a newly
proposed CF. Mechanical checks establish executability, the Verifier
assesses domain support, and the acceptance policy decides whether the
guidance reaches the frozen Solver.

The Verifier receives the original question and options, edited input,
CF operation and claimed effect, Proposer rationale, and retrieved
evidence. It does not receive reference answers, post-CF Solver answers,
correction/harm labels, or test scores. It returns an evidence-support
judgment rather than a benchmark answer.

\subsubsection{Model-specific retrieval indexes}

Within each dataset, all three configurations use the same knowledge
texts, independently encoded with their assigned embedding model
(Table~\ref{tab:mcq-embedding-models}). Vectors are L2-normalized and
stored in separate Chroma indexes; documents, IDs, and metadata remain
aligned. Queries follow the corresponding format: Qwen uses an instruction
with \texttt{Query:}, Nemotron uses \texttt{query:}/\texttt{passage:},
and EmbeddingGemma uses its task and document prefixes. Shared texts keep
the source knowledge comparable, while each index uses its own embedding
space; the retrieved passages need not be identical.

\begin{table}[!htbp]
\centering
\begin{minipage}{0.88\linewidth}
\centering
\caption{\textbf{Embedding models for MCQ retrieval.}}
\label{tab:mcq-embedding-models}
\small
\setlength{\tabcolsep}{3pt}
\renewcommand{\arraystretch}{1.10}
\begin{tabularx}{\linewidth}{@{}>{\raggedright\arraybackslash}p{0.22\linewidth}>{\raggedright\arraybackslash}Xr@{}}
\toprule
Base model & Embedding checkpoint & Dim. \\
\midrule
Qwen2.5-7B & \path{Qwen/Qwen3-Embedding-0.6B} & 1,024 \\
Llama3-8B & \path{nvidia/llama-nemotron-embed-1b-v2} & 2,048 \\
MedGemma-4B & \path{google/embeddinggemma-300m} & 768 \\
\bottomrule
\end{tabularx}
\end{minipage}
\end{table}

\subsubsection{Knowledge bases}

\textbf{Medical knowledge.}
DiagnosisArena uses 39 collections and 84,572 chunks: 36 topical PubMed
Central collections plus DailyMed, Surviving Sepsis Campaign guidelines,
and American College of Radiology Appropriateness Criteria. Topics span
major clinical areas. Llama and MedGemma re-encode the same source
passages, IDs, and metadata without adding benchmark answers or Solver
outputs. Table~\ref{tab:mcq-index-sizes} reports counts per model-specific
index; these are the MCQ indexes, not the earlier IHM/LOS configuration.

\begin{table}[!htbp]
\centering
\begin{minipage}{0.76\linewidth}
\centering
\caption{\textbf{Knowledge-base size per MCQ index.}}
\label{tab:mcq-index-sizes}
\small
\setlength{\tabcolsep}{3pt}
\renewcommand{\arraystretch}{1.10}
\begin{tabular*}{\linewidth}{@{\extracolsep{\fill}}lrr@{}}
\toprule
Knowledge corpus & Collections & Chunks \\
\midrule
DiagnosisArena medical & 39 & 84,572 \\
MMMU Business & 5 & 16,302 \\
\bottomrule
\end{tabular*}
\end{minipage}
\end{table}

\textbf{Business textbooks.}
Six OpenStax textbooks cover financial and managerial accounting,
economics, finance, management, and marketing. The accounting books share
one collection. Fixed PDFs are stored with SHA-256 hashes and converted
using \texttt{pdftotext -layout}. Deterministic cleaning removes front
matter, assessment and answer-key pages, and repeated page boilerplate.
Text is split into approximately 900-character chunks with 150-character
overlap, without joining across removed-page gaps.

To reduce benchmark overlap, question and option text from all 967
Business items is compared with the corpus using normalized exact
12-word n-grams. Twenty-four overlapping chunks are removed. Benchmark
text is used only for exclusion; reference answers and Solver outputs
are not added to the corpus. Table~\ref{tab:mcq-business-chunks} reports
the retained text shared by the three indexes.

\begin{table}[!htbp]
\centering
\begin{minipage}{0.68\linewidth}
\centering
\caption{\textbf{Business corpus after cleaning.}}
\label{tab:mcq-business-chunks}
\small
\setlength{\tabcolsep}{3pt}
\renewcommand{\arraystretch}{1.10}
\begin{tabular*}{\linewidth}{@{\extracolsep{\fill}}lr@{}}
\toprule
Business collection & Chunks \\
\midrule
\texttt{business\_accounting} & 4,189 \\
\texttt{business\_economics} & 4,767 \\
\texttt{business\_finance} & 2,789 \\
\texttt{business\_management} & 2,182 \\
\texttt{business\_marketing} & 2,375 \\
\midrule
Total & 16,302 \\
\bottomrule
\end{tabular*}
\end{minipage}
\end{table}

\subsubsection{Query construction and retrieval}

\textbf{Medical retrieval.}
Four queries cover the original condition, the intervention's diagnostic
meaning, the claimed outcome or preservation, and the decisive feature.
A topic router selects collections by disease, organ system, drug, or
imaging topic. The base configuration retrieves up to four chunks per
query and collection, deduplicates them, and supplies at most 24 passages
with source IDs to the Verifier.

\textbf{Business retrieval.}
The question's subject selects one collection. Queries cover the required
concept or formula, the edit's effect, the claimed conclusion, and the
decisive quantity. Retrieval applies a distance threshold of 0.75,
deduplicates by chunk ID, and selects up to three qualifying passages
with the smallest distances. The displayed score $1/(1+d)$ ranks stored
retrieval distances $d$; it is not a calibrated probability.

\subsubsection{Verifier judgments and safeguards}

Each configuration uses its corresponding frozen base model in a Verifier
call at temperature zero. It assesses the edited condition, intervention,
and claimed consequence. Medical \texttt{PASS}/\texttt{SUPPORT} indicates
relevant support; \texttt{FLAG}/\texttt{INSUFFICIENT} indicates partial,
indirect, or non-unique support; \texttt{REJECT}/\texttt{CONTRADICT}
indicates contradiction. Missing evidence produces uncertainty, not
automatic contradiction.

Business verification also checks formulas, variable roles, units, time
bases, aggregation levels, and linked totals or ratios. A text edit can
be executable yet violate an accounting identity. \texttt{PASS} requires
support for the condition, intervention, and outcome; contradiction gives
\texttt{REJECT}, and incomplete or non-unique support gives \texttt{FLAG}.
Unparseable final outputs are recorded as \texttt{ERROR}. Deterministic
formatting repairs do not change verdicts or business claims; at most one
strict-JSON retry is permitted, with raw responses retained.

\subsubsection{Acceptance and reproducibility}

A verdict does not overwrite the Solver answer. Each acceptance policy is
selected on validation and frozen for test. It may require support, veto
an explicit rejection, or ignore the RAG verdict; a non-adopted proposal
leaves the initial answer unchanged. Corpus and embedding manifests,
source and retrieved-passage IDs, raw judgments, and retrieval summaries
make the decisions traceable. The policy results in
Section~\ref{app:cf_acceptance_policies} evaluate how this evidence signal
is used. This separates an executable edit from a domain-supported claim
and from a proposal selected for downstream use.
\FloatBarrier

\section{Extended RQ2 Analyses}
\label{app:rq2-extended}

\subsection{Robustness across base models}
\begin{table*}[htbp]
\centering
\small
\caption{\textbf{Extended robustness and baseline comparisons (RQ2).} Metrics follow Figure~\ref{tab:rq2_robustness}, with additional clinical baselines. LOS is regression. A dash marks a matched result still in progress. Section~\ref{app:rq2-mcq-results} specifies the MCQ acceptance policies, retrieval baselines, and Business output processing.}
\label{tab:app-rq2-robustness-full}

\setlength{\tabcolsep}{3.5pt}
\begin{tabular}{lllrrrrr}
\toprule
Dataset / task & Metric & Base model & Zero-shot & RAG & GSEM & MILM & \textbf{Ours} \\
\midrule
\multirow{3}{*}{MIMIC-IV / IHM} & \multirow{3}{*}{AUROC} & Qwen2.5-7B  & 0.7000 & -- & -- & 0.7235 & \textbf{0.7553} \\
 & & Llama3-8B   & 0.7025 & -- & -- & 0.7089 & \textbf{0.7466} \\
 & & MedGemma-4B & 0.7117 & -- & -- & 0.7157 & \textbf{0.7374} \\
\addlinespace[2pt]
\multirow{3}{*}{MIMIC-IV / LOS} & \multirow{3}{*}{MAE $\downarrow$} & Qwen2.5-7B  & 69.76 & -- & -- & 59.77 & \textbf{58.60} \\
 & & Llama3-8B   & 59.42 & -- & -- & 65.26 & \textbf{55.20} \\
 & & MedGemma-4B & 56.29 & -- & -- & 60.14 & \textbf{55.42} \\
\addlinespace[2pt]
\multirow{3}{*}{DiagnosisArena} & \multirow{3}{*}{Acc. (\%)} & Qwen2.5-7B  & 34.8 & 26.8 & 26.8 & -- & \textbf{66.7} \\
 & & Llama3-8B   & 30.4 & 29.7 & 31.2 & -- & \textbf{61.6} \\
 & & MedGemma-4B & 37.7 & 33.3 & 33.3 & -- & \textbf{63.0} \\
\addlinespace[2pt]
\multirow{3}{*}{HoVer} & \multirow{3}{*}{Acc. (\%)} & Qwen2.5-7B  & 54.45 & -- & -- & -- & \textbf{61.92} \\
 & & Llama3-8B   & 59.00 & -- & -- & -- & \textbf{63.22} \\
 & & MedGemma-4B & 57.07 & -- & -- & -- & \textbf{58.86} \\
\addlinespace[2pt]
\multirow{3}{*}{MMMU Business} & \multirow{3}{*}{Acc. (\%)} & Qwen2.5-7B  & 61.4 & 53.8$^{\dagger}$ & 47.6 & -- & \textbf{63.4} \\
 & & Llama3-8B   & 42.1 & 42.8$^{\ddagger}$ & 26.9 & -- & \textbf{51.7} \\
 & & MedGemma-4B & 35.9 & 37.2$^{\ddagger}$ & 33.1 & -- & \textbf{47.6} \\
\midrule
\multicolumn{3}{@{}l}{FuseMoE (no LLM)} & \multicolumn{5}{l}{IHM AUROC 0.7246; LOS MAE 61.33} \\
\bottomrule
\end{tabular}
\end{table*}

\subsection{Proposer--Solver decomposition}
The SFT and SFT+GRPO rows use the same configurations as
Table~\ref{tab:app-rq1-cfit-full}; the heuristic row adds a training-free
comparison.

\begin{table}[htbp]
\centering
\small
\setlength{\tabcolsep}{4pt}
\begin{tabular}{llrrr}
\toprule
Solver & Proposer context & Accuracy & AUROC & AUPRC \\
\midrule
Qwen & None (zero-shot) & 0.5839 & 0.7000 & 0.2252 \\
Qwen & Heuristic CF & 0.5839 & 0.6968 & 0.2225 \\
Qwen & SFT Proposer & 0.7372 & 0.7543 & 0.2985 \\
Qwen & SFT + GRPO & \textbf{0.8417} & \textbf{0.7553} & \textbf{0.3237} \\
Qwen SFT & None & \textbf{0.8786} & 0.7584 & 0.3128 \\
Qwen SFT & Solver-specific SFT & 0.8630 & \textbf{0.7595} & \textbf{0.3212} \\
\bottomrule
\end{tabular}
\caption{Full-test IHM decomposition of Proposer instruction tuning,
outcome-guided RL, and direct Solver fine-tuning. Comparisons are controlled
within each Solver pair.}
\label{tab:app-rq2-decomposition}
\end{table}

\subsection{CF supervision scale and operation distributions}
\label{app:cf-supervision-analysis}
\begin{figure*}[htbp]
\centering
\includegraphics[width=0.88\textwidth]{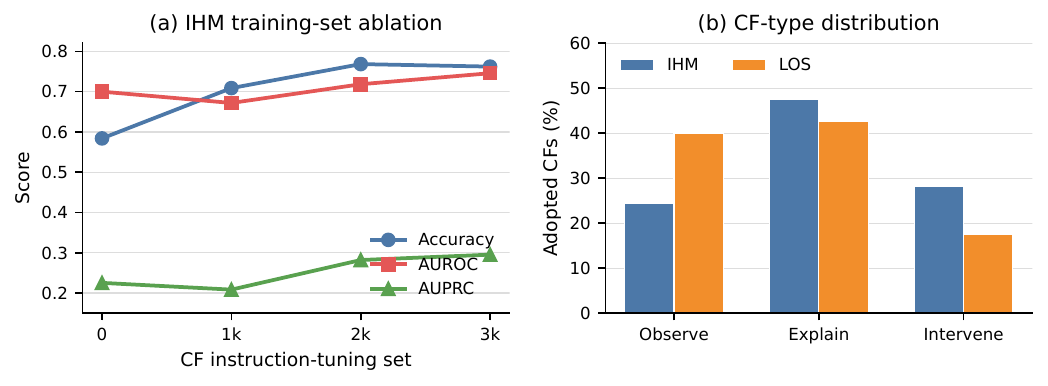}
\caption{Analysis of CF supervision. (a) IHM performance with zero-shot inference and CF Proposers trained on 1k, 2k, and 3k examples. (b) Adopted CFs by unified edit-level type (Appendix~\ref{app:cf-unified-types}) differ across tasks. IHM excludes 1,460 \texttt{no\_cf} outputs ($n=9{,}011$ actionable CFs); LOS reports its full evaluation ($n=14{,}825$).}
\label{fig:app-cf-supervision-analysis}
\end{figure*}

Figure~\ref{fig:app-cf-supervision-analysis}(a) reports the IHM data-scale ablation. One thousand examples primarily teach the output behavior, while 2k and 3k examples provide enough corrective and preservation boundaries to improve accuracy and ranking metrics jointly. Figure~\ref{fig:app-cf-supervision-analysis}(b) reports adopted CFs by unified edit-level type across tasks.

\subsection{Unified CF types across datasets}
\label{app:cf-unified-types}
Each dataset exposes CF operations in its own output contract: clinical XML
names semantic operations (Appendix~\ref{app:cf-data}), whereas MCQ tasks
name executable span edits (Table~\ref{tab:mcq-edit-contract}).
Table~\ref{tab:cf-unified-types} maps both onto the edit-level categories
used in Figure~\ref{tab:rq1_cfit_main}b. These categories describe the
operation, while Observe, Explain, and Intervene describe its causal
purpose. A cause swap adds an alternative explanation for fixed evidence
(Explain), so it counts as Add. Withholding or adding evidence, including
changing when a finding is observed, tests Observe; changing a treatment
or support action tests Intervene. Span edits test Observe and are reported
as Mixed when they combine different atomic operations. Clinical CFs use
one local edit. The mapping uses operation labels, and \texttt{no\_cf}
abstentions are excluded.

\begin{table}[!htbp]
\centering
\small
\caption{\textbf{Unified edit-level CF types and their dataset-specific operations.}}
\label{tab:cf-unified-types}
\setlength{\tabcolsep}{4pt}
\begin{tabularx}{\linewidth}{@{}lX>{\raggedright\arraybackslash}p{0.24\linewidth}>{\raggedright\arraybackslash}p{0.18\linewidth}@{}}
\toprule
Edit type & What changes & Clinical (MIMIC-IV) & Span edit (MCQ) \\
\midrule
Remove & An evidence item or channel is withheld & \texttt{modality\_ablation} & \texttt{delete} \\
\addlinespace[2pt]
Add & Hypothetical evidence or an alternative explanation is introduced & \texttt{evidence\_addition}, \texttt{cause\_swap} & \texttt{add} \\
\addlinespace[2pt]
Replace & A finding, observation time, treatment, or support state is substituted & \texttt{action\_change}, \texttt{temporal\_shift} & \texttt{replace} \\
\addlinespace[2pt]
Mixed & Several atomic edits of different types & -- (single edit) & combination \\
\bottomrule
\end{tabularx}
\end{table}

\subsection{Cross-model IHM configurations}
Table~\ref{tab:app-base-transfer} gives the full IHM metric vectors for the
configurations in Table~\ref{tab:app-rq2-robustness-full}, with a separately
trained Proposer for each base model. Its Qwen row matches the
SFT row in Table~\ref{tab:app-rq1-cfit-full}. Qwen and Llama each use 10,471
cases in both runs; MedGemma uses 10,336 zero-shot and 10,299 CF cases.
Figure~\ref{fig:counterfactual-motivation}(c) instead reports an earlier Qwen
Proposer configuration trained on the same 3,000 CFs: accuracy 0.7217, AUROC 0.7518,
and AUPRC 0.2881 on 10,470 CF cases, compared with 10,471 zero-shot cases.

\begin{table*}[htbp]
\centering
\small
\setlength{\tabcolsep}{4pt}
\begin{tabular}{llrrr}
\toprule
Base model & Context & Accuracy & AUROC & AUPRC \\
\midrule
Qwen2.5-7B & None / learned CF & 0.5839 / \textbf{0.7372} & 0.7000 / \textbf{0.7543} & 0.2252 / \textbf{0.2985} \\
Llama3-8B & None / learned CF & 0.6065 / \textbf{0.7424} & 0.7025 / \textbf{0.7466} & 0.2268 / \textbf{0.2830} \\
MedGemma-4B & None / learned CF & 0.7620 / \textbf{0.8027} & 0.7117 / \textbf{0.7374} & 0.2344 / \textbf{0.2906} \\
\bottomrule
\end{tabular}
\caption{Full IHM metrics for Table~\ref{tab:app-rq2-robustness-full}.}
\label{tab:app-base-transfer}
\end{table*}

\subsection{Cross-Task and Cross-Model MCQ Results}
\label{app:rq2-mcq-results}

\textbf{Comparison settings.}
Experiment~8 compares zero-shot inference, retrieval without evolution,
GSEM, and the full CF-guided system. DiagnosisArena uses static Top-1
CF-memory retrieval. On Business, $\dagger$ marks Qwen's Top-3 textbook
retrieval and $\ddagger$ marks Llama and MedGemma's legacy RAG results.
These baseline contexts are distinct from the full system's RAG Verifier
(Section~\ref{app:mcq-rag-verifier}). GSEM is an external baseline, not
a within-framework removal of CF generation.
Table~\ref{tab:app-rq2-cross-task} uses the recorded runs plotted in
Figure~\ref{tab:rq2_robustness}, not seed averages. Business accuracy
compares complete systems, including model-specific numerical
recalculation or output recovery. Corrected and Harmed measure the
subsequent paired CF revisions within each run.

\textbf{DiagnosisArena acceptance policies.}
The Qwen result (66.7\%, 47 corrections, three harms) uses the
validation-selected ``Ignore RAG signal'' policy in
Table~\ref{tab:app-verifier-policy}: at least one candidate vote and
reliability at least 0.96, with mechanical checks active. RAG verdicts do
not determine adoption in this configuration. Llama and MedGemma use
validation-calibrated gates requiring an executable \texttt{CROSSED}
proposal, a reliability threshold (approximately 0.8785 and 0.8175,
respectively), agreement between the revised Solver answer and the
Proposer answer, and calibrated confidence changes of at least $-0.05$
and $-0.15$. These policies do not require a positive RAG verdict.
Section~\ref{app:cf_acceptance_policies} compares alternative uses of
Verifier feedback and their correction--preservation tradeoffs.

\begin{table}[!htbp]
\centering
\begin{minipage}{0.94\linewidth}
\centering
\caption{\textbf{MCQ results across tasks and base models.}}
\label{tab:app-rq2-cross-task}
\small
\setlength{\tabcolsep}{2.5pt}
\renewcommand{\arraystretch}{1.10}
\begin{tabular*}{\linewidth}{@{\extracolsep{\fill}}lrrrrrr@{}}
\toprule
& \multicolumn{4}{c}{Accuracy (\%)} & \multicolumn{2}{c}{Answer changes} \\
\cmidrule(lr){2-5}\cmidrule(l){6-7}
Base model & Zero-shot & RAG & GSEM & \textbf{Ours} & Corrected & Harmed \\
\midrule
\multicolumn{7}{@{}l}{\textit{DiagnosisArena} ($n=138$)}\\
Qwen2.5-7B & 34.8 & 26.8 & 26.8 & \textbf{66.7} & 47 & 3 \\
Llama3-8B & 30.4 & 29.7 & 31.2 & \textbf{61.6} & 43 & 0 \\
MedGemma-4B & 37.7 & 33.3 & 33.3 & \textbf{63.0} & 35 & 3 \\
\addlinespace[4pt]
\multicolumn{7}{@{}l}{\textit{MMMU Business} ($n=145$)}\\
Qwen2.5-7B & 61.4 & $53.8^{\dagger}$ & 47.6 & \textbf{63.4} & 4 & 1 \\
Llama3-8B & 42.1 & $42.8^{\ddagger}$ & 26.9 & \textbf{51.7} & 23 & 13 \\
MedGemma-4B & 35.9 & $37.2^{\ddagger}$ & 33.1 & \textbf{47.6} & 3 & 1 \\
\bottomrule
\end{tabular*}
\end{minipage}
\end{table}

\textbf{Paired gains across models.}
The full system has the highest recorded accuracy among the listed
methods in each dataset--model setting. Table~\ref{tab:mcq-paired-starts}
uses each full-system run's initial predictions, which can differ from
a separate zero-shot run because of output processing. Every row satisfies
$B+C-H=N_{\mathrm{correct,final}}$, with gains computed from unrounded
counts. DiagnosisArena gains range from 23.2 to 31.9 percentage points,
with 35--47 corrections and 0--3 harms. Llama corrects 43 errors and
preserves all initially correct answers in this run.

\begin{table}[!htbp]
\centering
\begin{minipage}{0.86\linewidth}
\centering
\caption{\textbf{Paired correction and preservation results.} Initial correct is measured after each run's output processing; Gain measures the subsequent CF revision step. Separate zero-shot accuracies appear in Table~\ref{tab:app-rq2-cross-task}.}
\label{tab:mcq-paired-starts}
\small
\setlength{\tabcolsep}{3pt}
\renewcommand{\arraystretch}{1.10}
\begin{tabular*}{\linewidth}{@{\extracolsep{\fill}}lrrrrr@{}}
\toprule
Base model & \shortstack{Initial\\correct} & \shortstack{Final\\correct} & Corrected & Harmed & \shortstack{Gain\\(pp)} \\
\midrule
\multicolumn{6}{@{}l}{\textit{DiagnosisArena} ($n=138$)}\\
Qwen2.5-7B & 48 & 92 & 47 & 3 & +31.9 \\
Llama3-8B & 42 & 85 & 43 & 0 & +31.2 \\
MedGemma-4B & 55 & 87 & 35 & 3 & +23.2 \\
\addlinespace[4pt]
\multicolumn{6}{@{}l}{\textit{MMMU Business} ($n=145$)}\\
Qwen2.5-7B & 89 & 92 & 4 & 1 & +2.1 \\
Llama3-8B & 65 & 75 & 23 & 13 & +6.9 \\
MedGemma-4B & 67 & 69 & 3 & 1 & +1.4 \\
\bottomrule
\end{tabular*}
\end{minipage}
\end{table}

Business system-level gains over the separate zero-shot runs are 2.1,
9.7, and 11.7 percentage points for Qwen, Llama, and MedGemma, computed
from unrounded counts. The corresponding paired CF gains are 2.1, 6.9,
and 1.4 points. For MedGemma, output recovery raises the starting count
from 52 to 67 correct answers; CF revision then yields $67+3-1=69$.
Thus the main comparison reports total system gains, while
Table~\ref{tab:mcq-paired-starts} measures CF's incremental effect.
Across all six paired runs, corrections exceed harms, with the largest
gains on DiagnosisArena.
\FloatBarrier

\subsection{Adopted CF Operations and Evidence-Use Errors}
\label{app:mcq-evidence-use}

Experiment~9 examines the DiagnosisArena consensus configuration from
Experiment~4. Of 41 adopted interventions, 39 correct a wrong answer,
two switch between wrong options, and none harms an initially correct
answer (Table~\ref{tab:app-exp9-adopted}). The 95.1\% correction share
uses adopted interventions as its denominator, not all test questions.
Replace is the most frequent primary operation; these frequencies
describe adoption rather than an operation-specific ablation.

\begin{table}[!htbp]
\centering
\begin{minipage}{0.78\linewidth}
\centering
\caption{\textbf{Adopted DiagnosisArena CF interventions.}}
\label{tab:app-exp9-adopted}
\small
\setlength{\tabcolsep}{3pt}
\renewcommand{\arraystretch}{1.10}
\begin{tabular*}{\linewidth}{@{\extracolsep{\fill}}lrr@{}}
\toprule
Category & Count & Share (\%) \\
\midrule
\multicolumn{3}{@{}l}{\textit{Adopted outcomes ($n=41$)}}\\
Wrong $\rightarrow$ correct & 39 & 95.1 \\
Wrong $\rightarrow$ different wrong & 2 & 4.9 \\
Correct $\rightarrow$ wrong & 0 & 0.0 \\
\addlinespace[4pt]
\multicolumn{3}{@{}l}{\textit{Primary CF operation ($n=41$)}}\\
Replace & 19 & 46.3 \\
Add & 11 & 26.8 \\
Mixed & 11 & 26.8 \\
\bottomrule
\end{tabular*}
\end{minipage}
\end{table}

\begin{figure}[!htbp]
\centering
\includegraphics[width=0.86\linewidth]{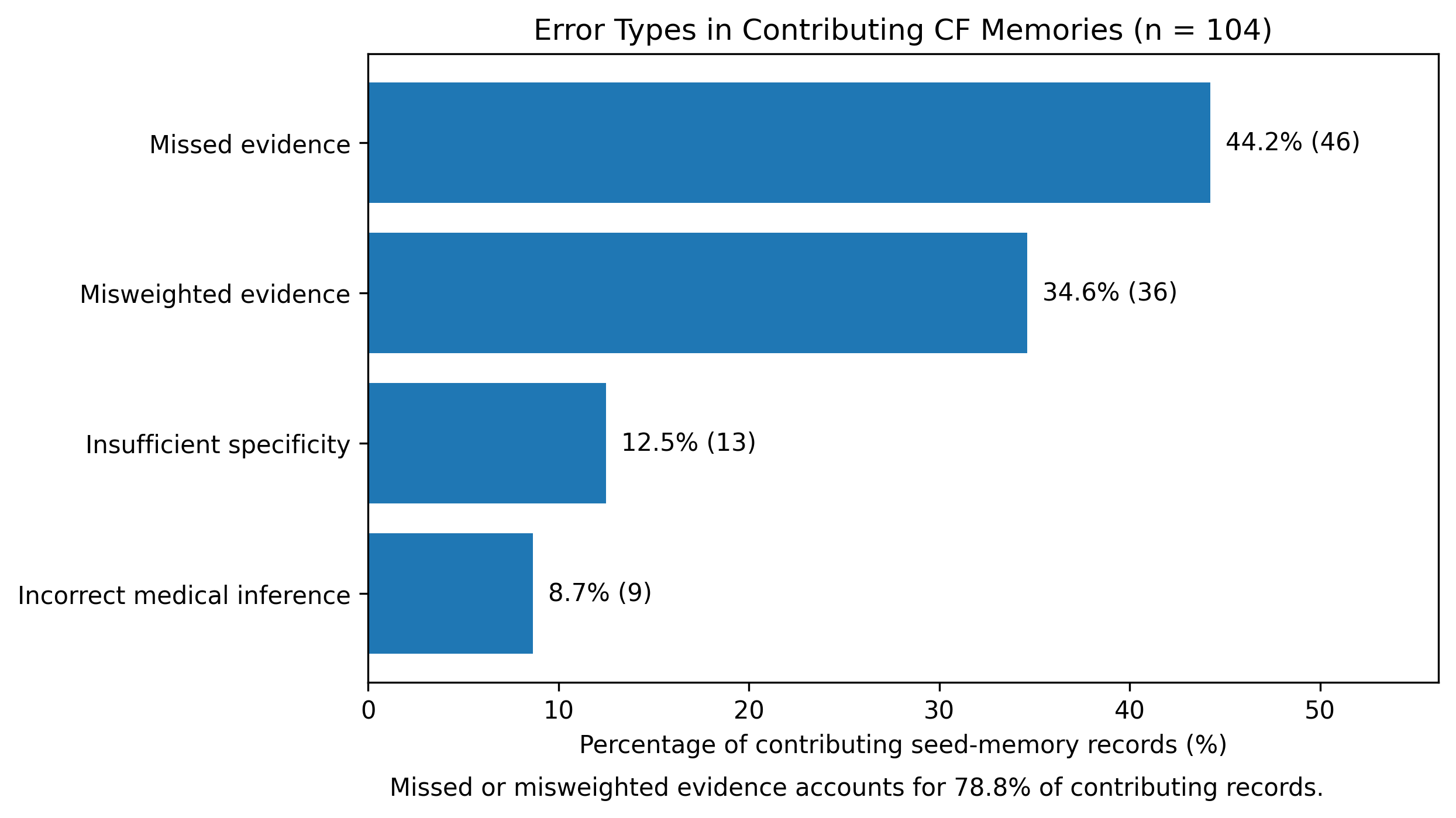}
\caption{\textbf{Error types in contributing CF memories.}
The 104 DiagnosisArena seed-memory records include 44.2\% missed evidence
and 34.6\% misweighted evidence.}
\label{fig:app-exp9-error-types}
\end{figure}

\textbf{Evidence-use errors dominate contributing memories.}
Among 104 contributing seed-memory records, 46 concern missed evidence
and 36 concern misweighted evidence: together, 78.8\%. The remaining
13 concern insufficient specificity and nine incorrect medical inference
(Figure~\ref{fig:app-exp9-error-types}). Multiple seed memories may
contribute to one intervention; the 104 records are not independent test
questions. This distribution is consistent with the intended role of CF
guidance: prompting the Solver to reconsider which findings support its
answer and how much weight they deserve.

\FloatBarrier

\subsection{CF Acceptance Policies Using RAG Verifier Feedback}
\label{app:cf_acceptance_policies}

Experiment~7 compares how RAG judgments enter CF-acceptance decisions
with Qwen2.5-7B on DiagnosisArena. All policies share the three-seed
candidate pool, retrieved evidence, frozen Solver replays, and Verifier
judgments. Mechanical checks remain active. Each policy selects its own
reliability, vote, and margin conditions on 137 validation questions,
then is frozen for the 138-question test. This compares complete
acceptance policies rather than a single Verifier on/off switch.

Policies ignore RAG judgments, veto an explicit \texttt{REJECT}, or accept
\texttt{PASS}/\texttt{FLAG} verdicts in combination with their other
conditions. \texttt{FLAG} denotes incomplete support, not full
verification. The two named single-vote variants use the shared candidate
pool, not newly trained single-seed Proposers. In
Table~\ref{tab:app-verifier-policy}, accuracy and preservation are
percentages. Adopted, Corrected, Harmed, and Preserved refer to test;
preservation is measured among the 48 initially correct answers. Each
row is one policy result, not an average across independent policy runs.

\begin{table}[!htbp]
\centering
\begin{minipage}{0.94\linewidth}
\centering
\caption{\textbf{RAG-Verifier acceptance policies on DiagnosisArena.}}
\label{tab:app-verifier-policy}
\small
\setlength{\tabcolsep}{2pt}
\renewcommand{\arraystretch}{1.10}
\begin{tabularx}{\linewidth}{@{}>{\raggedright\arraybackslash}Xrrrrrr@{}}
\toprule
Policy & \shortstack{Val.\\acc.} & \shortstack{Test\\acc.} & Adopted & Corrected & Harmed & \shortstack{Preserved\\(\%)} \\
\midrule
Ignore RAG signal & 67.9 & \textbf{66.7} & 55 & 47 & 3 & 93.8 \\
\addlinespace[2pt]
Veto explicit REJECT & 56.9 & 52.9 & 32 & 27 & 2 & 95.8 \\
\addlinespace[2pt]
Positive-only (PASS/FLAG) & 52.6 & 48.6 & 21 & 19 & \textbf{0} & \textbf{100.0} \\
\addlinespace[2pt]
Single-vote + REJECT veto & 67.2 & \textbf{66.7} & 51 & 46 & 2 & 95.8 \\
\addlinespace[2pt]
Single-vote + positive support & 66.4 & 64.5 & 44 & 41 & \textbf{0} & \textbf{100.0} \\
\bottomrule
\end{tabularx}
\end{minipage}
\end{table}

\textbf{Correction and preservation.}
The ``Ignore RAG signal'' row supplies the Qwen headline result in
Figure~\ref{tab:rq2_robustness} and Table~\ref{tab:app-rq2-cross-task}.
The single-vote REJECT-veto policy matches the 66.7\% accuracy of ignoring
RAG judgments. Its 46 corrections and two harms yield the same net gain
of 44 correct answers as 47 corrections and three harms, with 51 rather
than 55 adopted interventions. The single-vote positive-support policy
reaches 64.5\%, corrects 41 errors, and preserves all initially correct
answers in this test run. These outcomes show how validation-selected
policies can retain substantial correction benefits while choosing a
more conservative revision pattern. The other policies provide the full
comparison: benefits depend on how the verdict and selection conditions
are combined, not simply on whether a Verifier is present.
\FloatBarrier

\section{Comparison with Direct Solver Fine-Tuning}
\label{app:solver_comparisons}

Experiment~11 asks where to apply supervision: directly to the Solver,
or to a Proposer guiding a frozen Solver. Data-matched Solver-SFT uses
the same training question IDs as the corresponding Proposer. On
DiagnosisArena, the baseline uses answer-only SFT. On Business, Qwen and
MedGemma use SFT adapters evaluated with their zero-shot reasoning and
parsing protocol; Llama uses reasoning-aligned SFT on the same 400
training questions. The Ours column repeats the best recorded complete
Experiment~8 runs, not seed means. Business results include numerical
recalculation or output recovery, so this remains a system-level
comparison (Table~\ref{tab:frozen_vs_sft}).

\begin{table}[!htbp]
\centering
\begin{minipage}{0.82\linewidth}
\centering
\caption{\textbf{Comparison with direct Solver fine-tuning (accuracy, \%).}}
\label{tab:frozen_vs_sft}
\small
\setlength{\tabcolsep}{3pt}
\renewcommand{\arraystretch}{1.10}
\begin{tabular*}{\linewidth}{@{\extracolsep{\fill}}lrrr@{}}
\toprule
Base model & Zero-shot & Solver-SFT & Ours \\
\midrule
\multicolumn{4}{@{}l}{\textit{DiagnosisArena} ($n=138$)}\\
Qwen2.5-7B & 34.8 & \textbf{74.6} & 66.7 \\
Llama3-8B & 30.4 & \textbf{79.0} & 61.6 \\
MedGemma-4B & 37.7 & \textbf{70.3} & 63.0 \\
\addlinespace[4pt]
\multicolumn{4}{@{}l}{\textit{MMMU Business} ($n=145$)}\\
Qwen2.5-7B & 61.4 & 62.1 & \textbf{63.4} \\
Llama3-8B & 42.1 & 49.7 & \textbf{51.7} \\
MedGemma-4B & 35.9 & 44.1 & \textbf{47.6} \\
\bottomrule
\end{tabular*}
\end{minipage}
\end{table}

Both approaches improve over the listed zero-shot results. Solver-SFT
is higher on DiagnosisArena, while the frozen-Solver systems are higher
on Business. The comparison supports a practical alternative to updating
the decision-making model: learning CF guidance can yield useful
full-system gains while leaving Solver parameters unchanged.
\FloatBarrier

\section{Additional Results for RQ3}
\label{app:rq3}

\subsection{Patient-specific CF curriculum}
\label{app:rq3-curriculum}

We define three task-difficulty levels by prediction horizon: from the same
first-24-hour record, the frozen Solver predicts mortality within 24 (Easy), 72
(Medium), or 168 hours (Hard). On a fixed paired cohort of 2,000 patients,
zero-shot AUROC decreases from $0.7697$ to $0.7305$ and $0.7142$, establishing
increasingly difficult tasks without changing the observed input. Within each
task, we further partition patients into Easy, Medium, and Hard tertiles using
only the frozen zero-shot confidence. Starting with no CF, we progressively make
one patient-specific CF available to the Easy, then Medium, and finally Hard
subgroup; patients not yet covered retain their exact zero-shot probability.
Thus each round expands the applicable CF context without updating either model
or revealing additional patient observations.
Because mortality prevalence rises with the horizon, we compare tasks mainly
with AUROC and balanced accuracy, which do not depend on prevalence, and report
Macro-AUPRC and Macro-F1 as secondary metrics.

\begin{figure}[ht]
  \centering
  \includegraphics[width=0.6\linewidth]{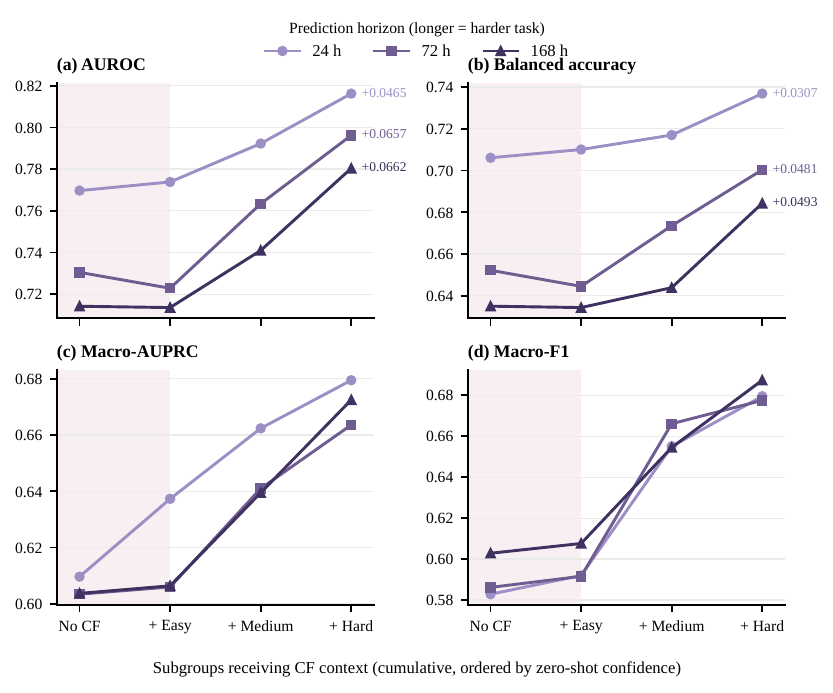}
  \caption{\textbf{Difficulty-ordered accumulation of CF context.} Patients are
  grouped by frozen zero-shot confidence, and CF context is admitted cumulatively
  from Easy to Medium to Hard. Adding Easy context alone has little effect,
  whereas the larger gains emerge after Medium and Hard cases are included,
  especially for the longer-horizon tasks. Darker lines are longer (harder)
  horizons; the shaded region marks the Easy-only round, and labels in (a) and
  (b) give the final gain over no CF.}
  \label{fig:cf-subgroup-curriculum}
\end{figure}

Figure~\ref{fig:cf-subgroup-curriculum} shows little or slightly negative change
when only Easy experience is added, followed by consistent gains as Medium and
Hard experience becomes available. At the final round, AUROC reaches $0.8162$,
$0.7962$, and $0.7804$ for Easy, Medium, and Hard prediction, corresponding to
gains of $0.0465$, $0.0657$, and $0.0662$. Balanced-accuracy gains similarly
increase from $0.0307$ on Easy to $0.0481$ and $0.0493$ on Medium and Hard. The
ordered increase shows that additional CF context contributes little to already
easy decisions but becomes increasingly useful for harder patient subgroups and
harder tasks.
The AUROC gap between the 24-hour and 168-hour tasks shrinks from $0.0555$
without CF to $0.0358$ with full CF. The Medium and Hard rounds account for at
least 91\% of the final AUROC gain at every horizon, and at 72 hours CF on the
Easy subgroup alone lowers AUROC by $0.0076$ and balanced accuracy by $0.0077$.
Full CF context also improves Macro-AUPRC ($+0.060$ to $+0.070$) and Macro-F1
($+0.084$ to $+0.097$) at every horizon, but these prevalence-dependent metrics
do not show larger gains at longer horizons.

\subsection{Held-out CF memory transfer}
\label{app:rq3-memory}

\textbf{Setup.} All cases that already have a CF form the memory pool (517
source cases); the remaining cases are split into a validation set (177) and a
frozen test set (710). Train, validation, and test share no record IDs, no test
case contributes its own CF or memory, and no memory source appears in the test
set. The test set contains 480 MMMU and 230 MMMU-Pro questions (207 Easy, 327
Medium, 176 Hard). We compare an empty bank with four banks built from source
cases of increasing Solver confidence: Low (189 memories), Low+Medium (338), All
(517), and High-only (179). Each non-empty condition retrieves the top-1 memory.
The Solver, prompt, parser, and safety gate are frozen before the test run, and
the Low bank, best on validation, is the main configuration; the other banks are
ablations.

\begin{figure}[ht]
  \centering
  \includegraphics[width=0.6\linewidth]{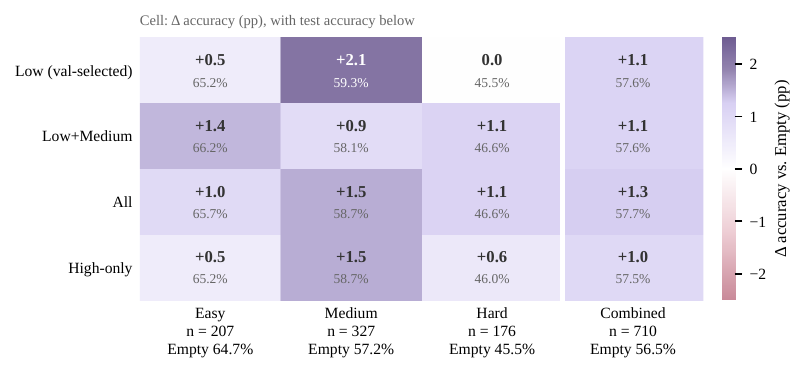}
  \caption{\textbf{Held-out accuracy with CF memory.} Each cell shows the change
  in test accuracy over the empty bank (bold) and the resulting test accuracy
  (below).}
  \label{fig:memory-difficulty}
\end{figure}

\textbf{Accuracy by difficulty.} All four banks improve overall accuracy over the
empty bank (56.5\%), reaching 57.5--57.7\% and finishing within two test
questions of each other (Figure~\ref{fig:memory-difficulty}). Growing the bank
from 189 to 517 memories therefore adds little. The main bank gains most on
Medium questions (+2.1 pp) and leaves Hard questions unchanged; the larger banks
add two correct Hard answers (+1.1 pp).

\begin{figure}[ht]
  \centering
  \includegraphics[width=0.7\linewidth]{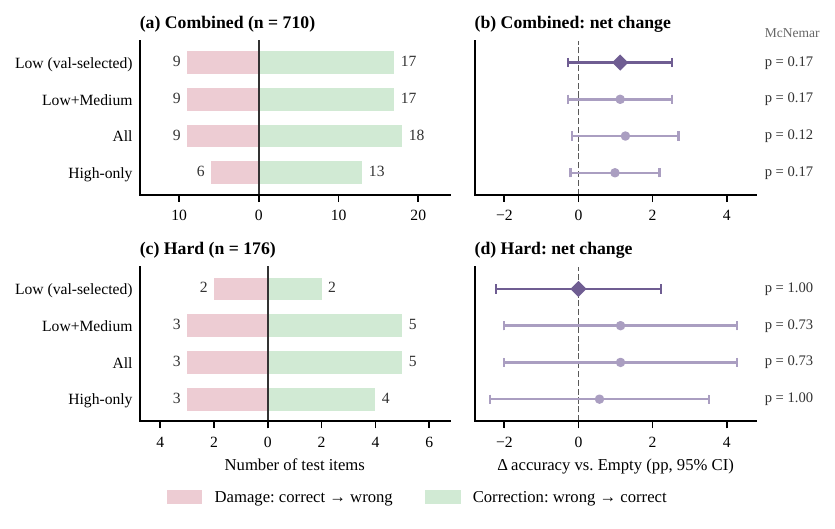}
  \caption{\textbf{Memory fixes and breaks answers, and the net gain is
  uncertain.} (a, c) Corrections (wrong $\rightarrow$ correct) and damage
  (correct $\rightarrow$ wrong) relative to the empty bank. (b, d) Net change in
  test accuracy with 95\% paired confidence intervals and exact McNemar
  $p$-values.}
  \label{fig:memory-flips-ci}
\end{figure}

\textbf{Corrections, damage, and uncertainty.} The net gain hides changes in
both directions (Figure~\ref{fig:memory-flips-ci}). Each bank fixes roughly two
wrong answers for every correct answer it breaks; the main bank fixes 17 and
breaks 9, and the High-only bank makes fewer changes in both directions (13 and
6). All confidence intervals include zero, both overall and on Hard questions.

\begin{figure}[ht]
  \centering
  \includegraphics[width=0.7\linewidth]{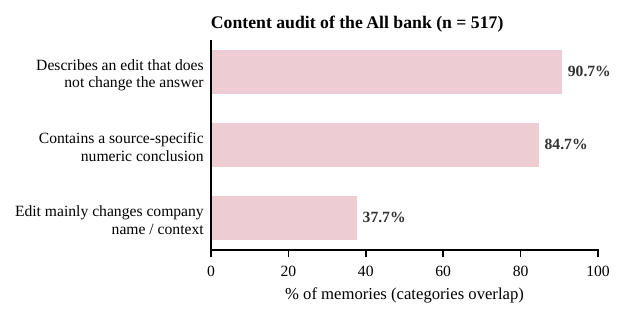}
  \caption{\textbf{Content audit of the full memory bank.} Share of the 517
  memories with each property; a memory can fall into several categories.}
  \label{fig:memory-audit}
\end{figure}

\textbf{Memory content.} An audit of the full bank
(Figure~\ref{fig:memory-audit}) shows that 90.7\% of memories describe an edit
that did not change the answer, 84.7\% contain numeric conclusions tied to their
source question, and 37.7\% mainly change a company name or context. Such
memories give the Solver little it can reuse on a new question, so the
bottleneck is the information in each memory rather than the number of
memories.

\textbf{Limitations.} Memory sources were selected because a CF already existed,
not at random, and all memories come from MMMU while the test set also includes
MMMU-Pro.

\section{Analysis of Memory Evolution}
\label{app:memory-evolution}
We specialize the principle of high-confidence policy improvement \citep{thomas2015highconfidence} to the memory trajectory in Section~\ref{sec:memory-evolution}. The result compares complete memory snapshots, including their effects on retrieval and Solver decisions. Passing the existing gate for an individual CF does not imply that the resulting bank improves accuracy.

\paragraph{Assumptions for Proposition 3.}
Let $\mathcal H$ contain the initial model configuration, evolution data, and all choices and randomness used to construct $\mathcal M_0,\ldots,\mathcal M_R$. Successive rounds may depend on earlier memories and feedback from the evolution pool. The model weights remain fixed, and each round adds notes from disjoint new cases. Proposer optimization precedes these rounds, and the teacher is used only for the initial instruction-tuning warm-up. Fix the round budget $R$ and the $S$ target strata before evaluation. For each stratum $s$, let $\mathcal D_s$ be its fixed task distribution and $\mathcal V_s=\{Z_{si}\}_{i=1}^{n_s}$ an i.i.d. sample independent of $\mathcal H$. Evaluation inputs, labels, and scores do not determine memory construction, retrieval settings, or subsequent evolution choices. Memory is read-only during evaluation; the context budget and the retrieval, selection, and ordering rules are fixed before evaluation. Use deterministic decoding or a fixed stochastic protocol with independent randomness across sampled units; all expectations below include that randomness. Correlated records from one patient are not separate independent units. Different strata may share units, and rounds reuse the same cases; independence is required within each stratum, not across rounds or strata.

For $Z=(x,y)$, write $C_r(Z)=\mathbf{1}\{\hat y(x;\mathcal M_r)=y\}$ and, conditional on $\mathcal H$, define
\begin{equation}
\begin{aligned}
  a_{r,s}&=\mathbb E_{Z\sim\mathcal D_s}[C_r(Z)],\\
  \Delta_{r,s}&=a_{r,s}-a_{r-1,s},\\
  \widehat\Delta_{r,s}
    &=\frac{1}{n_s}\sum_{i=1}^{n_s}
      \bigl[C_r(Z_{si})-C_{r-1}(Z_{si})\bigr].
\end{aligned}
\label{eq:memory-paired-gain}
\end{equation}
Under the specified pairing, let $b_{r,s}=\Pr(C_{r-1}=0,C_r=1)$ and $h_{r,s}=\Pr(C_{r-1}=1,C_r=0)$ within stratum $s$. As in Equation~\ref{eq:correction-preservation}, $\Delta_{r,s}=b_{r,s}-h_{r,s}$ and $\widehat\Delta_{r,s}=\widehat b_{r,s}-\widehat h_{r,s}$. The bound therefore asks whether the observed excess of corrections over harmful reversals exceeds sampling uncertainty.

\paragraph{Proposition 3: simultaneous improvement bounds.}
Under these assumptions, for any prespecified $\delta\in(0,1)$, the bounds in Equation~\ref{eq:memory-evolution-lcb} satisfy
\begin{equation}
  \Pr\!\left(
    \Delta_{r,s}\ge L_{r,s}
    \text{ for all }r=1,\ldots,R,\ s=1,\ldots,S
  \right)\ge1-\delta.
  \label{eq:memory-simultaneous-bound}
\end{equation}
On this event, any transition with $\min_s L_{r,s}>0$ improves population accuracy in every target stratum. If the condition holds at every round, every stratum's population accuracy increases along the trajectory. No such increase is assumed in establishing the bound.

\noindent\textbf{Proof.}
Condition on $\mathcal H$, which fixes the memory trajectory independently of the evaluation samples. For any $(r,s)$, the differences $D_{si}=C_r(Z_{si})-C_{r-1}(Z_{si})$ are independent across $i$, have mean $\Delta_{r,s}$, and lie in $[-1,1]$. Hoeffding's inequality \citep{hoeffding1963probability} gives
\begin{equation}
  \Pr\!\left(\widehat\Delta_{r,s}-\Delta_{r,s}>\epsilon
    \mid\mathcal H\right)
  \le \exp\!\left(-\frac{n_s\epsilon^2}{2}\right).
\end{equation}
Set $\epsilon=\sqrt{2\log(RS/\delta)/n_s}$. The probability that any particular lower bound exceeds its true gain is at most $\delta/(RS)$. A union bound over the $RS$ comparisons gives Equation~\ref{eq:memory-simultaneous-bound} conditional on $\mathcal H$. Averaging over $\mathcal H$ gives the unconditional statement. The union bound does not require independent comparisons, so reusing evaluation cases across rounds, or patients across strata, does not invalidate it. Positivity of the relevant lower bounds proves the improvement statements.

\paragraph{Acceptance interpretation and scope.}
The result supplies a sufficient acceptance criterion for a prespecified memory transition. It does not establish that our CF admission gate enforces that criterion, nor that a failed certificate implies an update is harmful. Reusing the frozen evaluation sets is compatible with the proof only while the memory trajectory remains independent of them. If their scores trigger rollback, change retrieval, or guide later proposals, the resulting trajectory requires a different argument, for example fresh independent validation batches with round-wise error budgets summing to $\delta$. Data used for such selection are validation data, and final testing must remain separate.

The reported RQ3 experiment (Figure~\ref{fig:cf-subgroup-curriculum}) cumulatively enables patient-specific CF context for confidence-defined Easy, Medium, and Hard subgroups on a fixed paired cohort. Both models remain frozen, and patients not yet covered retain their zero-shot probabilities. This tests expanding CF coverage without an acceptance or rollback step. Proposition~3 applies to memory trajectories satisfying the independence assumptions above; its accuracy bound does not certify the reported AUROC or balanced-accuracy gains.

Each bound applies to its own $\mathcal D_s$. A hard or rare stratum may differ from the evolution pool and still be covered when it has its own independent evaluation sample. Improvement on an easier stratum alone does not certify improvement on a harder one. The theorem concerns mean correctness under fixed stratum distributions; it does not guarantee zero harm to individual cases, valid hypothetical outcomes, or improvement in AUROC, F1, or other non-additive metrics. An uncertainty band across model seeds is also distinct from this bound on sampling error across task instances.

\end{document}